\documentclass[runningheads]{llncs}

\usepackage{eccv}

\usepackage{eccvabbrv}

\usepackage{graphicx}
\usepackage{booktabs}

\usepackage[accsupp]{axessibility}

\usepackage{hyperref}

\usepackage{orcidlink}
\usepackage{lipsum}

\usepackage{lipsum}
\newcommand\blfootnote[1]{%
	\begingroup
	\renewcommand\thefootnote{}\footnote{#1}%
	\addtocounter{footnote}{-1}%
	\endgroup
}

\usepackage{marvosym}
\usepackage{multirow}
\usepackage{colortbl}
\usepackage{xcolor}
\usepackage{wrapfig}
\definecolor{mypink}{rgb}{.99,.91,.95}
\definecolor{backgroundcolor}{RGB}{199, 238, 206}

\begin{document}


\title{Camera Calibration using a Collimator System} 

\author{Shunkun Liang\orcidlink{0000-0002-2020-2414}\and
		Banglei Guan\textsuperscript{\Letter} \orcidlink{0000-0003-2123-0182} \and
		Zhenbao Yu \orcidlink{0000-0003-0274-2684}  \and
		Pengju Sun\orcidlink{0009-0001-0075-0667}  \and
		Yang Shang \orcidlink{0000-0002-5836-5016}}

\authorrunning{S.~Liang, B.~Guan et al.}

\institute{College of Aerospace Science and Engineering, National University of Defense Technology, China.
\blfootnote{\Letter~Corresponding author. \  \email{guanbanglei12@nudt.edu.cn}}}
\maketitle

\begin{abstract}
	Camera calibration is a crucial step in photogrammetry and 3D vision applications. In practical scenarios with a long working distance to cover a wide area, target-based calibration methods become complicated and inflexible due to site limitations. This paper introduces a novel camera calibration method using a collimator system, which can provide a reliable and controllable calibration environment for cameras with varying working distances. Based on the optical geometry of the collimator system, we prove that the relative motion between the target and camera conforms to the spherical motion model, reducing the original 6DOF relative motion to 3DOF pure rotation motion. Furthermore, a closed-form solver for multiple views and a minimal solver for two views are proposed for camera calibration. The performance of our method is evaluated in both synthetic and real-world experiments, which verify the feasibility of calibration using the collimator system and demonstrate that our method is superior to the state-of-the-art methods. Demo code is available at \url{https://github.com/LiangSK98/CollimatorCalibration}.
\end{abstract}

\section{Introduction}
\label{sec:intro}
Geometric camera calibration determines the mapping between 3D rays in space and 2D image pixels.Various vision applications, such as simultaneous localization and mapping(SLAM)\cite{Campos2021}, structure from motion(SfM)\cite{Colmap2016}, and pose estimation\cite{Guan2023,GuanTCYB2021}, rely on camera parameters as inputs. The close-range calibration methods using a planar target represent the prevailing techniques in 3D vision applications\cite{Brown1971,Sturm1999,Zhang2000,Lochman2021,Thomas2020}. In some special outdoor scenarios, such as long-range depth estimation\cite{Zhang2020} and monitoring of wind turbine blades\cite{Guan2022}, the cameras achieve a long working distance and a large field of view. The traditional target-based calibration methods \cite{Tsai1987,Zhang2000,Bouguet2004} may face challenges, such as needing large targets, moving or arranging targets over a wide area, and dealing with varying and unstable lighting conditions. While constructing a 3D calibration field can offer precise calibration data \cite{Wang2015,Xiao2010,Shang2013}, its operational complexity and limited adaptability present significant challenges. In addition, the collimator is commonly used to generate calibration targets\cite{Clarke1998}. Traditional collimator-based methods need to measure the direction of collimated rays using a goniometer or theodolite, which makes the calibration process challenging, and measurement errors directly impact the accuracy\cite{Yuan2019,Yuan2021,Wu2007}. 
\begin{figure}[tbp]
	\centering
	\includegraphics[width=\linewidth]{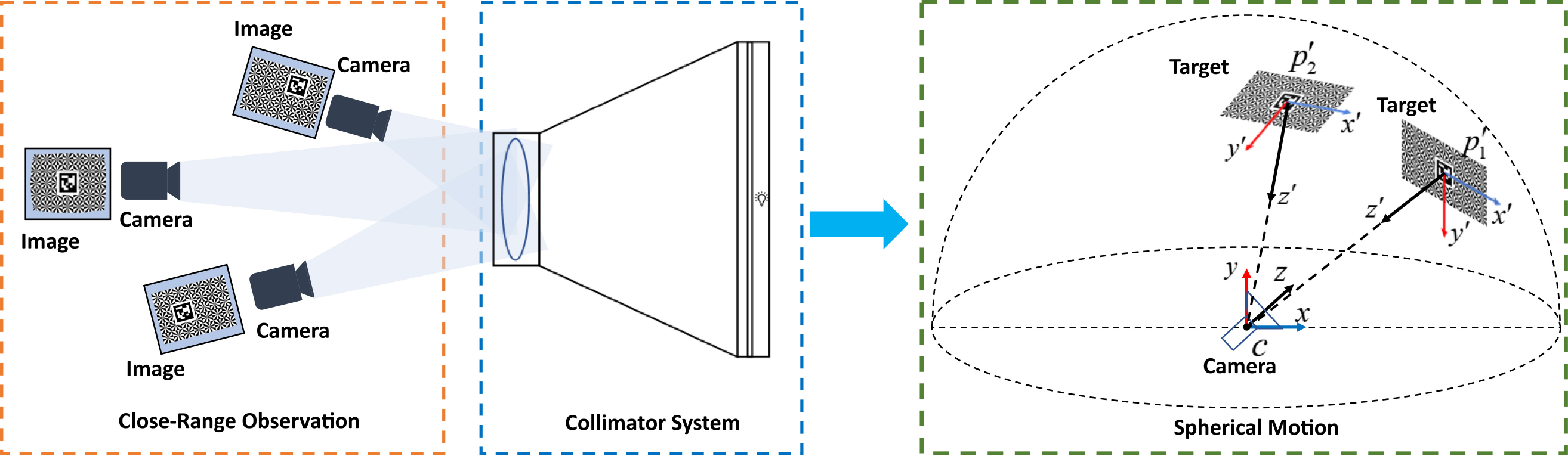}
	\caption{Diagram of the collimator-based calibration method. From the inherent optical geometric properties of the collimator, the relative motion between the target and camera can be proved to conform to the spherical motion model.}
	\label{fig:coll_calib}
\end{figure}

In this paper, we design a novel collimator system with a structured calibration pattern on its reticle, and a novel calibration algorithm based on the collimator images is proposed. Our method only requires the camera to observe the pattern from our collimator system at a few (at least two) different orientations. \cref{fig:coll_calib} illustrates this method. We briefly analyze the geometric optical property of the collimator and illustrate the angle invariance of any point pairs on the pattern. By leveraging angle invariance, we prove that the relative motion between the calibration target and the camera conforms to a spherical motion model. This constraint effectively reduces the original 6DOF general motion to the 3DOF pure rotation motion. Furthermore, we propose a novel calibration method using the collimator system. Compared with the target-based methods, more geometric constraints can be derived from the collimator, leading to higher accuracy. Compared to traditional collimator-based methods, our method is more flexible as it eliminates the need for direction measurement. 

Our primary contributions can be summarized as follows:
(1) We propose a novel camera calibration method using a collimator system. The entire calibration process can be implemented with multiple close-range observations and does not require the known direction of collimated rays. 
(2) We prove that the relative motion between the calibration target and the camera conforms to the spherical motion model. This constraint effectively reduces the number of motion parameters to be estimated.
(3) A closed-form solver for multiple views and a minimal solver for two views are proposed based on the spherical motion constraint. Moreover, the degenerate case of the proposed solvers is proved.

\section{Related Work}
\label{sec:Related}
Camera calibration is a crucial subject in 3D vision and continues to be actively researched. The literature on camera calibration introduces a wide range of algorithms and toolboxes \cite{Brown1971,Opencv2000,Kannala2006,Zhang2000,Sturm1999,Bouguet2004,Lochman2021,Thomas2020}. To ensure the highest accuracy for calibration, targets with known structures are usually used for camera calibration. Various calibration targets have been proposed, including 3D targets \cite{Tsai1987}, 2D targets\cite{Heikkila2000,Ha2017} and 1D targets\cite{Zhang2004}. In some practical applications where no calibration target is available, the camera calibration can be performed by establishing the correspondence between images during the movement of the camera, which is known as self-calibration\cite{Berlin1992, Pollefeys1999, Nister2005, Hartley1994, Herrera2016, Gustavo2017}. Among these options, the target-based close-range calibration method is widely used.

In outdoor long-range measurement applications\cite{Zhang2020,Guan2022}, the camera has a long working distance to measure the long-distance targets. Thus, the close-range calibration methods are impractical due to the site limitation. A commonly employed approach involves constructing a calibration field, which serves as a known structure for the calibration\cite{Liu2015,Hu2022,Wang2015,Shang2013,Xiao2010,Oniga2018}. Oniga \etal \cite{Oniga2018} lay markers on buildings to calibrate cameras mounted on UAV. Shang \etal \cite{Shang2013} proposed a high-precision calibration method by placing several markers near the ground. Additionally, Klaus \etal \cite{Klaus2004} proposed using stars as distant targets to estimate camera parameters. 

The collimator is also commonly used to provide 2D-3D correspondence for camera calibration in the laboratory\cite{Bauer2008, Yuan2019, Hieronymus2012, Wu2007}. Previous collimator-based methods proposed to use goniometer or theodolite to obtain the direction of collimated rays and then estimate the camera parameters\cite{Yuan2021,Wu2007}. The collimator-based method offers both accuracy and flexibility in the limited space, but a high-precision goniometer or theodolite is expensive. To address this, Yuan \etal \cite{Yuan2019} designed a multiple pinhole mask for the collimator to generate the collimated rays with known directions. However, the accuracy of calibration is sensitive to machining errors. Bauer \etal \cite{Bauer2008} proposed a calibration method using a diffractive optical element (DOE). The angle between the laser beam and the camera can then be estimated to determine the camera parameters.

\section{Collimator System}
\label{sec:Collimator}
This section introduces the designed collimator system and shows that the equivalent relative motion between the calibration target and the camera conforms to a spherical motion model.
\subsection{Scheme of Collimator System}
\label{sec:Scheme}
The collimator is an optical instrument that can produce collimated rays and plays an important role when adjusting other optical instruments. A collimator consists of essential components such as a light source, a reticle, and an optical lens, as illustrated in \cref{fig:Collimator_Camera}. The reticle is a thin glass located on the focal plane of the collimator. The generation of collimated rays can be regarded as the inverse process of imaging a target at infinity onto the focal plane. Specifically, a point on the lens's focal plane will generate collimated rays, which can be regarded as the target at infinity. Therefore, cameras with varying focal lengths can consistently and accurately observe the target through the collimator at a close range.

In this paper, we design a collimator system for camera calibration. A star-based pattern, proposed by Thomas \etal \cite{Thomas2020}, is attached to the reticle of the collimator as the calibration target. The star-based pattern offers richer gradient information for improving feature extraction accuracy. Similar to the plane-based calibration method \cite{Sturm1999,Zhang2000}, the star-based pattern provides 2D-3D correspondences for camera calibration. Our collimator system is portable with a size of about $200mm \times 170mm \times 300mm$. The collimator system is designed to provide a reliable and controllable calibration environment for some special scenarios like inappropriate illumination, limited space, and long-range cameras. However, printed targets face challenges in these scenarios. When calibrating a camera, we only need to capture the target from multiple orientations through the lens of the collimator, which is as easy as the printed targets method.

\subsection{Spherical Motion Model}
\label{sec:Motion Model}
\begin{wrapfigure}{r}{0.5\linewidth}
	\centering
	\includegraphics[width=0.9\linewidth]{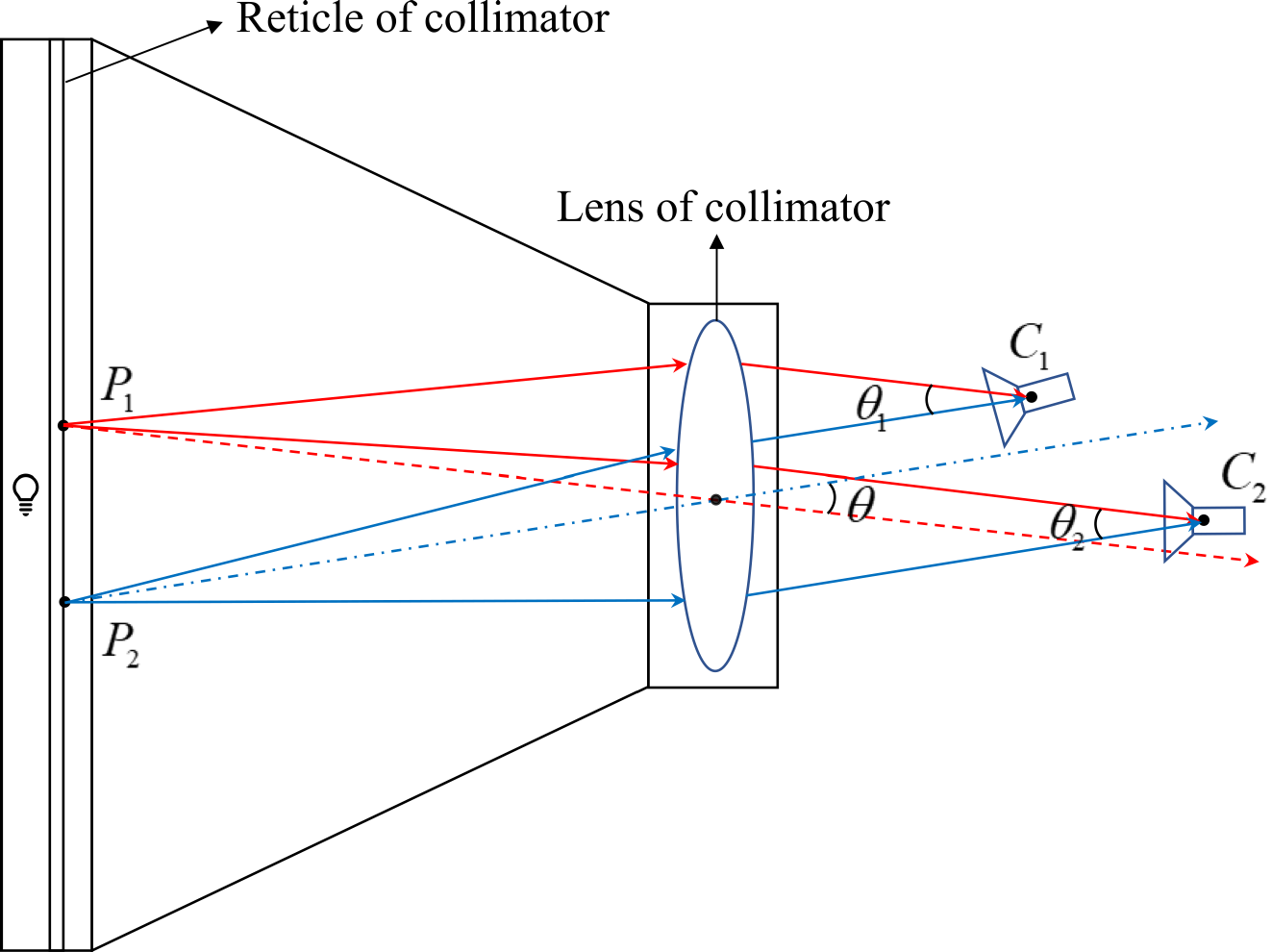}
	\caption{Diagram of collimator system. The geometric property of the collimator system leads to the angle invariance of any point pair on the reticle.}
	\label{fig:Collimator_Camera}
\end{wrapfigure}
In this section, we prove that the relative motion between the calibration target and the camera conforms to the spherical motion model. Firstly, we briefly analyze the geometric property of the collimator. As shown in \cref{fig:Collimator_Camera}, let us note two arbitrary points $P_1$ and $P_2$ on the reticle of the collimator. Rays emitted from $P_1$ and $P_2$ pass through the collimator lens, generating two parallel beams observed by cameras denoted as $C_1$ and $C_2$ with arbitrary poses. The propagation routes of the two beams are visually distinguished by red and blue lines in \cref{fig:Collimator_Camera}. Obviously, the angle between two lines observed by the two cameras is equal, that is, $\theta_1 \equiv \theta_2 \equiv \theta$. We can obtain a property that the two rays observed by the cameras always maintain a constant angle, regardless of the poses of observing cameras. We refer to this property as the angle invariance. 

Next, we show that this setup, where a camera moves around a collimator system and captures images, is theoretically equivalent to the capture setting where the camera is fixed, but the images are moving around the camera spherically. This can be derived from the angle invariance. For convenience in introducing our derivation process, we treat the camera as a fixed reference coordinate, and assume that the calibration target moves relative to the camera in space. The angle invariance allows us to infer that when the target performs a specific motion in space, the angular distance between any point pair on the target always remains constant. The angular distance means the angle between two vectors from the origin (camera optical center) to two points. By arbitrarily selecting a pair of points $\mathbf{P}_i$ and $\mathbf{P}_j$ on the target and transforming them, we can establish the following equation
\begin{equation}
	\angle\left(\mathbf{P}_i , \mathbf{P}_j \right) = \angle\left(\mathbf{R}\left(\mathbf{P}_i+\mathbf{t}\right), \mathbf{R}\left(\mathbf{P}_j +\mathbf{t}\right)\right),
	 \label{eq:angPiPj}
\end{equation}
where $\mathbf{R}$ is rotation matrix and $\mathbf{t}$ is translation vector. From the rotation invariance\cite{Wang2022} that a pair of 3D points are jointly rotated about the origin, their angular distance remains constant. We have $\angle\left(\mathbf{R}\left(\mathbf{P}_i+\mathbf{t}\right), \mathbf{R}\left(\mathbf{P}_j +\mathbf{t}\right)\right) = \angle\left(\mathbf{P}_i+\mathbf{t}, \mathbf{P}_j +\mathbf{t}\right)$. Thus, the angle invariance is maintained when the calibration target is arbitrarily rotated about the origin. To make $\angle(\mathbf{P}_i, \mathbf{P}_j)) = \angle(\mathbf{P}_i+\mathbf{t}, \mathbf{P}_j +\mathbf{t})$ hold for any pair of points on the target, the translation vector must be a $\mathbf{0}$ vector ($\mathbf{t}=[0,0,0]^T$). The proof is given in the \textit{supplementary material}. Therefore, the calibration target performs pure rotation motion with respect to the origin (\ie camera optical center). Such pure rotational motion is a typical spherical motion. 

The spherical motion can be visualized in \cref{fig:coll_calib}. In the spherical motion model, the calibration target exhibits only rotational motion with a fixed distance relative to the camera. We establish a local coordinate on the target, where the $z^{\prime}$-axis aligns with the ray between the camera coordinate system $c$ and calibration target coordinate system $p^{\prime}$. Inspired by \cite{Ventura2016}, the relative pose between the calibration target and camera can be mathematically represented as 
\begin{equation}
	\mathbf{T}_{cp^\prime} = \left( {\begin{array}{*{20}{c}}
			{\mathbf{R}_{cp^\prime}}&\mathbf{t}_{cp^\prime}\\
			{\mathbf{0}_{1 \times 3}}&1
	\end{array}} \right),
	\label{eq:Tcpp}
\end{equation}
where $\mathbf{R}_{cp^\prime} \in SO(3) $, $\mathbf{t}_{cp^\prime} = {(0,0,-r)^T}$ and $r$ is the radius of spherical motion. We usually establish the origin of the coordinate system $p$ on the upper left corner of the calibration pattern. Therefore, there is a relative translation $\mathbf{t}_{p^{\prime} p} = (x,y,0)$ between the coordinate systems $p^{\prime}$ and $p$. Furthermore, the transformation matrix between camera $c$ and calibration target $p$ is expressed as follows:
\begin{equation}
	\mathbf{T}_{pc} = \left[\left( {\begin{array}{*{20}{c}}
			{\mathbf{R}_{cp^\prime}}&\mathbf{t}_{cp^\prime}\\
			{\mathbf{0}_{1 \times 3}}&1
	\end{array}} \right) \left( {\begin{array}{*{20}{c}}
			{\mathbf{I}_{3\times3}}&\mathbf{t}_{p^\prime p}\\
			{\mathbf{0}_{1 \times 3}}&1
	\end{array}} \right)\right]^{-1} =
	\left( {\begin{array}{*{20}{c}}
			{{\mathbf{R}_{pc}}}&{ - {\mathbf{R}_{pc}}{\mathbf{t}_{cp}}}\\
			{\mathbf{0}_{1 \times 3}}&1
	\end{array}} \right),
	\label{eq:Tpc}
\end{equation}
where $\mathbf{R}_{pc} = \mathbf{R}_{cp^\prime} ^ {T}$ and  ${\mathbf{t}_{cp}}\!=\!\mathbf{t}_{p^{\prime} p}\!+\!\mathbf{t}_{cp^\prime}\!=\!(x,y,-r)^T$. $\mathbf{t}_{cp}$ determines the position of the camera in the calibration target coordinate system. Note that $\mathbf{t}_{cp}$ is a fixed value and has nothing to do with the pose of the calibration target. Given the constraint from the spherical motion, the original 6DOF general motion is limited to a 3DOF pure rotation motion.

\section{Calibration Method}
\label{sec:Calibration}
The spherical motion constraint effectively reduces the number of motion parameters to be estimated. In this section, we show how to calibrate camera parameters using a set of collimator images. A closed-form solver for multiple views and a minimal solver for two views are proposed.
\subsection{Geometric Constraints}

We use the well-known pinhole model to model cameras. The mapping from a 3D point $\mathbf{P} = (X,Y,Z)^T$ to an image point $\mathbf{p} = (u,v)^T$ is computed as
\begin{equation}
	s \widetilde{\mathbf{p}}=\mathbf{K}[\mathbf{R} \mid \mathbf{t}] \widetilde{\mathbf{P}} \text {, with } \mathbf{K}=\left[\begin{array}{ccc}
		f_{x} & \gamma & c_{x} \\
		0 & f_{y} & c_{y} \\
		0 & 0 & 1
	\end{array}\right],
	\label{eq:KRtP}
\end{equation}
where $s$ is an arbitrary scale factor, $(\mathbf{R}, \mathbf{t})$ is the camera pose with respect to the calibration target coordinate, and $\mathbf{K}$ is the camera intrinsic matrix. $\widetilde{\mathbf{p}} = [u,v,1]^T$ and $\widetilde{\mathbf{P}} = [X,Y,Z,1]^T$ denote the homogeneous coordinate forms of points $\mathbf{p}$ and $\mathbf{P}$, respectively. The intrinsic matrix $\mathbf{K}$ consists of five parameters: the focal length $(f_x,f_y)$, the principal point $(c_x,c_y)$ and skew factor $\gamma$. 

Without loss of generality, we assume the planar target is on the plane $Z = 0$ of the world coordinate system. The transformation between the image plane and calibration target plane can be constrained by a $3 \times 3$ homography matrix, denoted as $\mathbf{H}$. For perspective projection, the homography matrix $\mathbf{H}$ depends on the image pose $(\mathbf{R},\mathbf{t})$ and intrinsic matrix $\mathbf{K}$:
\begin{equation}
	\mathbf{H} = \left[ {\begin{array}{ccc}
			\mathbf{h}_1 & \mathbf{h}_2 & \mathbf{h}_3
	\end{array}} \right] = \lambda \mathbf{K} \left[ {\begin{array}{ccc}
			\mathbf{r}_1&\mathbf{r}_2&\mathbf{t}
	\end{array}} \right],
	\label{eq:H}
\end{equation}
where $\lambda$ is an arbitrary scalar, $\mathbf{r}_i$ is the $i$-th columns of $\mathbf{R}$. 
Given an image of the calibration target, the homography matrix $\mathbf{H}$ can be estimated using the plane point correspondences. Combining \cref{eq:Tpc} and \cref{eq:H}, we can get 
\begin{equation}
	\mathbf{M} = \left[ {\begin{array}{*{20}{c}}
			\mathbf{{r_1}}&\mathbf{{r_2}}& -\mathbf{R} \mathbf{t}_{cp}
	\end{array}} \right] = \frac{1}{\lambda} \mathbf{K}^{-1} \mathbf{H}.
\end{equation}
Using the knowledge that $\mathbf{R}$ is an orthogonal matrix, we have 
\begin{align}
	{\mathbf{H}^T}{\mathbf{K}^{ - T}}{\mathbf{K}^{ - 1}} \mathbf{H} = \lambda^2 \left[ {\begin{array}{ccc}
			1&0&{ - x}\\
			0&1&{ - y}\\
			{ - x}&{ - y}&{{{\left\| {{\mathbf{t}_{cp}}} \right\|}^2}}
	\end{array}} \right].
	\label{eq:HKKH}
\end{align}
There are 8 parameters to be estimated, 5 for intrinsic matrix and 3 for $\mathbf{t}_{cp} = (x,y,-r)^T$. For each given image, \cref{eq:HKKH} provides 5 independent constraints. Theoretically, a minimum of two images can completely solve all unknown parameters. In the next two subsections, we give a closed-form solver for $N(N>2)$ collimator images and a minimal solver for two collimator images. 

\subsection{Closed-form Solver for Multiple Views}
When there are more than two collimator images for calibration, we model the problem as a linear system and propose a closed-form solution. Suppose that the homography matrices $\mathbf{H}_i (i = 1,...,N)$ have been estimated for all images. The transformations between images $i$ and $j$ can be expressed as
\begin{equation}
	\mathbf{H}_j^{-1} \mathbf{H}_i = \frac{\lambda_i}{\lambda_j} \mathbf{M}_j^{-1} \mathbf{M}_i.
	\label{eq:HjHi}
\end{equation}
For the matrix $\mathbf{M}$ of each image, we can calculate the determinant as
\begin{equation}
	\begin{aligned}
		\left|\mathbf{M} \right| &= \left| \begin{array}{ccc} \mathbf{r}_1&\mathbf{r}_2&-\mathbf{R}\mathbf{t}_{cp} \end{array} \right| = \left| \begin{array}{ccc} \mathbf{r}_1 & \mathbf{r}_2 & -x\mathbf{r}_1-y\mathbf{r}_2+r\mathbf{r}_3 \end{array} \right| \\
		&= -x\left| \begin{array}{ccc} \mathbf{r}_1 & \mathbf{r}_2 & \mathbf{r}_1 \end{array} \right| 
		- y\left| \begin{array}{ccc} \mathbf{r}_1 & \mathbf{r}_2 & \mathbf{r}_2 \end{array} \right| 
		+ r\left| \begin{array}{ccc}
			\mathbf{r}_1 & \mathbf{r}_2 & \mathbf{r}_3 \end{array} \right| = r,
	\end{aligned}
\end{equation}
where $r$ is a non-zero fixed value. Using the fact that $\det(\mathbf{M}_j^{-1} \mathbf{M}_i) = 1$, we have $\lambda_{ij} = \lambda_i / \lambda_j = \sqrt[3]{|\mathbf{H}_j^{-1} \mathbf{H}_i|}$. Then, we choose the image with the most feature points observed as the base image. For convenience, here the first image is selected as the base image and we get $\lambda_{i1} = \sqrt[3]{|\mathbf{H}_j^{-1} \mathbf{H}_1|}$. 
By inverting \cref{eq:HKKH} of image $i$, we get
\begin{equation}
	\mathbf{H}_i^{-1} \mathbf{K} \mathbf{K}^T \mathbf{H}_i^{-T} = 
	\frac{1}{\lambda_i^2 r^2} \left[ {\begin{array}{*{20}{c}}
			{{r^2} + {x^2}}&{xy}&x\\
			{xy}&{{r^2} + {y^2}}&y\\
			x&y&1
	\end{array}} \right].
	\label{eq:HKKH2}
\end{equation}
Substitute $\lambda_i = \lambda_{i1} \lambda_1$ into \cref{eq:HKKH2}, we get
\begin{equation}
	\mathbf{H}_i^{-1} \mathbf{K} \mathbf{K}^T \mathbf{H}_i^{-T} =
	\frac{1}{\lambda_{i1}^2} \mathbf{A}, \text{ with }	\mathbf{A} = \left[ {\begin{array}{{ccc}}
			\frac{r^2+x^2}{(\lambda_1 r)^2}&\frac{xy}{(\lambda_1 r)^2}& \frac{x}{(\lambda_1 r)^2}\\
			\frac{xy}{(\lambda_1 r)^2}&\frac{r^2+y^2}{(\lambda_1 r)^2}& \frac{y}{(\lambda_1 r)^2}\\
			\frac{x}{(\lambda_1 r)^2}&\frac{y}{(\lambda_1 r)^2}& \frac{1}{(\lambda_1 r)^2}
	\end{array}} \right].
	\label{eq:HKKH3}
\end{equation}
Let us define a 6-vector $\mathbf{a}$ to represent matrix $\mathbf{A}$:
\begin{equation*}
	\mathbf{a} = (A_{11}, A_{12}, A_{13}, A_{22}, A_{23}, A_{33})^T,
\end{equation*}
where $A_{mn}$ is the value of $m$-th row and $n$-th column of matrix $\mathbf{A}$. Let
\begin{equation}
	\begin{aligned}
		\mathbf{W} =\left[\begin{array}{ccc}
			{c_x}^2 + {f_x}^2 + \gamma^2 & c_x c_y + f_y \gamma & c_x \\
			c_x c_y + f_y \gamma & {c_y}^2 + {f_y}^2 & c_y \\
			c_x & c_y & 1
		\end{array}\right].
	\end{aligned}
	\label{eq:W}
\end{equation}
Note that $\mathbf{W}$ is a symmetric matrix and can be defined by a 5-vector:
\begin{equation*}
	\mathbf{w} = (W_{11}, W_{12}, W_{13}, W_{22}, W_{23})^T, 
\end{equation*}
where $W_{mn}$ is the value of $m$-th row and $n$-th column of matrix $\mathbf{W}$. Obviously, \cref{eq:HKKH3} represents a linear system of equations involving the unknowns $[\mathbf{w}^T, \mathbf{a}^T]^T$. Combining \cref{eq:HKKH3,eq:W}, we can get the following equations for each image:
\begin{equation}
	\left[ {\begin{array}{*{20}{c}}
			{\mathbf{V}}&{ - {\lambda _{i1}^2} {\mathbf{I}_{5 \times 5}}}
	\end{array}} \right]\left[ {\begin{array}{*{20}{c}}
			\mathbf{w}\\ \mathbf{a}
	\end{array}} \right] = \mathbf{b},
	\label{eq:VI}
\end{equation}
with
\begin{equation*}
	\begin{aligned}
		&\mathbf{V} = [\mathbf{v}_{11}, \mathbf{v}_{12}, \mathbf{v}_{13}, 
		\mathbf{v}_{22}, \mathbf{v}_{23}]^T, \\
		&\mathbf{v}_{mn} =  {\left[h_{m1} h_{n1}, h_{m1} h_{n2}+h_{m2} h_{n1}, h_{m1} h_{n3}+h_{m3} h_{n1},\right.} \\
		&\left.  \qquad \ \ h_{m2} h_{n2}, h_{m2} h_{n3} + h_{m3} h_{n2}\right], \\
		&\mathbf{b} = - [h_{13}h_{13}, h_{13}h_{23}, h_{13}h_{33}, h_{23}h_{23}, h_{23}h_{33}]^T, 
	\end{aligned}
\end{equation*}
where $h_{mn}$ denotes the value of $m$-th row and $n$-th column of matrix $\mathbf{H}^{-1}$. Given $N(N>2)$ collimator images, we can stack $N$ such equations as \cref{eq:VI} together in matrix form:
\begin{equation}
	\mathbf{D} \left[ {\begin{array}{*{20}{c}}
			\mathbf{w}\\ \mathbf{a}
	\end{array}} \right] = \mathbf{b},
	\label{eq:Dwa}
\end{equation}
where $\mathbf{D}$ is a $5N\times11$ matrix. The closed-form solution of \cref{eq:Dwa} is given by
\begin{equation}
	\left[ {\begin{array}{*{20}{c}}
			\mathbf{w}\\ \mathbf{a}
	\end{array}} \right] =
	(\mathbf{D}^T \mathbf{D})^{-1}\mathbf{D}^T \mathbf{b}.
\end{equation}
Once $\mathbf{w}$ and $\mathbf{a}$ are estimated, we can calculate all camera intrinsic parameters and optical center $\mathbf{t}_{cp}$ by decomposing $\mathbf{W}$ and $\mathbf{A}$ respectively:
\begin{align*}
	c_x &= W_{13}, \quad c_y = W_{23}, \quad f_y = \sqrt{W_{22} - {c_y}^2}, \\
	\gamma &= (W_{12} - c_x c_y)/f_y, \quad	f_x = \sqrt{W_{11} - {c_x}^2 - {\gamma}^2}, \\
	x &= A_{13} / A_{33}, \quad y = A_{23} / A_{33}, \\	
	r &= -\sqrt{(A_{11}/A_{33}) - (A_{13}/A_{33})^2}.
\end{align*}
Subsequently, the rotation matrix of each image can be calculated with \cref{eq:H}:
\begin{equation}
	\mathbf{r}_1 = (1/\lambda ){\mathbf{K}^{ - 1}} \mathbf{h}_1, \quad
	\mathbf{r}_2 = (1/\lambda ){\mathbf{K}^{ - 1}}\mathbf{h}_2, \quad
	\mathbf{r}_3 = \mathbf{r}_1 \times \mathbf{r}_2,
\end{equation}
where $\lambda = (\left \| {\mathbf{K}^{ - 1}} \mathbf{h}_1 \right \| + \left \| {\mathbf{K}^{ - 1}} \mathbf{h}_2 \right \|)/2$. Finally, a singular value decomposition for re-orthogonalizing the rotation matrix  $\mathbf{R} = [\mathbf{r}_1, \mathbf{r}_2, \mathbf{r}_3]$ is required.

\subsection{Minimal Solver for Two Views}
When only two collimator images are used for calibration, the rank of the coefficient matrix in \cref{eq:Dwa} is determined to be 10. Solving rank-deficient equations presents a complex challenge. To address this issue, we construct a nonlinear equation system and propose a novel approach that utilizes the hidden variable technique\cite{Hartley2012} to estimate the unknowns in the minimum configuration.

Estimation of $\mathbf{W} = \mathbf{K}^{-T} \mathbf{K}^{-1}$ is the core of calibration. Note that $\mathbf{W}$ is a symmetric matrix, defined by a 6-vector $\mathbf{w} = (W_{11}, W_{12}, W_{13}, W_{22}, W_{23}, W_{33})^T$. We can extract the following 5 constraints from \cref{eq:HKKH}
\begin{align}
	\mathbf{v}_{12}^T \mathbf{w} &= 0 \label{eq:v12} \\
	(\mathbf{v}_{11}^T - \mathbf{v}_{22}^T) \mathbf{w} &= 0 \label{eq:v11}\\
	\mathbf{v}_{13}^T \mathbf{w} + x \mathbf{v}_{11}^T \mathbf{w} &=0 \label{eq:v13} \\
	\mathbf{v}_{23}^T \mathbf{w} + y \mathbf{v}_{11}^T \mathbf{w} &=0 \label{eq:v23} \\
	\mathbf{v}_{33}^T \mathbf{w} -  {\left\| {{\mathbf{t}_{cp}}} \right\|^2} \mathbf{v}_{11}^T \mathbf{w} &=0 \label{eq:v33}.
\end{align}
$\mathbf{v}_{mn}$ can be calculated as following:
\begin{equation*} 
	\begin{aligned}
		\mathbf{v}_{mn}= & {\left[h_{1m} h_{1n}, h_{1m} h_{2n}+h_{1n} h_{2m}, h_{1m} h_{3n} + h_{1n} h_{3m}\right.} \\
		& \left.h_{2m} h_{2n}, h_{2m} h_{3n}+h_{2n} h_{3m}, h_{3m}h_{3n}\right]^{T},
	\end{aligned}
\end{equation*}
where $h_{mn}$ denotes the value of $m$-th row and $n$-th column of matrix $\mathbf{H}$. Given only 2 collimator images, we can obtain 10 independent equations for 8 unknowns. The hidden variable technique \cite{Hartley2012} is employed to solve the above polynomial equations. Combining \cref{eq:v13,eq:v23,eq:v33}, we have
\begin{equation}
	(\mathbf{v}_{13}^T + \mathbf{v}_{23}^T + \mathbf{v}_{33}^T)\mathbf{w} + c \mathbf{v}_{11}^T \mathbf{w} =0,
	\label{eq:v1233}
\end{equation}
where $c = (x+y-{\left\| {{\mathbf{t}_{cp}}} \right\|^2})$. We first select constraints \cref{eq:v12,eq:v11,eq:v1233} to establish a new polynomial equation system. Next, we treat $c$ as a hidden variable, and the complete set of equations can be expressed in matrix form as
\begin{equation}
	\mathbf{C}(c) \mathbf{w}=0, \text{ with } \mathbf{C}(c) = \left[ {\begin{array}{*{20}{c}}
			{\mathbf{v}_{12}^T}\\
			{(\mathbf{v}_{11}^T - \mathbf{v}_{22}^T)}\\
			{	(\mathbf{v}_{13}^T + \mathbf{v}_{23}^T + \mathbf{v}_{33}^T) + c\mathbf{v}_{11}^T}
	\end{array}} \right],
\end{equation}
where the coefficient matrix $\mathbf{C}(c)$ contains the hidden variable $c$. Two images provide 6 equations for solving $c$, and $\mathbf{C}(c)$ is a $6 \times 6$ matrix. Since the equation system $\mathbf{C}(c) \mathbf{w}=0$ must have at least one nontrivial solution for $\mathbf{w}$, the determinant of $\mathbf{C}(c)$ is equal to zero. The equation, $\det (\mathbf{C}(c)) =0 $, is a polynomial of degree 2 in terms of $c$, which can be efficiently solved using a straightforward root-finding algorithm. Substituting the known value of $c$ into the constraint equations \cref{eq:v12,eq:v11,eq:v1233}, we obtain six constraint equations about $\mathbf{w}$. The least squares method is then employed to solve these linear equations and compute the optimal $\mathbf{w}$ up to scale. Once $\mathbf{w}$ is estimated, the intrinsic matrix $\mathbf{K}$ can be calculated using the Cholesky factorization\cite{Hartley2004}.

\subsection{Bundle Adjustment}

Bundle adjustment jointly refines the camera intrinsic $\mathbf{K}$, lens distortion coefficients $\mathbf{d}$, the rotation matrix of the $i$-th image $\mathbf{R}_i$ and camera optical center $\mathbf{t}_{cp}$. In cases of moderate lens distortion, it is typical to initialize the distortion coefficient to $\mathbf{0}$ and estimate them in the bundle adjustment step\cite{Triggs2000}. Given a set of $N$ images, each with $M$ corresponding 2D image points, we establish a cost function aimed at minimizing the re-projection error. The cost function is expressed as follows:
\begin{equation}
	\mathop {\min }\limits_\mathbf{X} \sum\limits_i^N {\sum\limits_j^M {\rho \left( {{{\left\| {\pi \left( {\mathbf{K}, \mathbf{d}, {\mathbf{R}_i},{\mathbf{t}_{cp}},{\mathbf{P}_{ij}}} \right) - {\mathbf{p}_{ij}}} \right\|}^2}} \right)} },
	\label{eq:BA}
\end{equation}
where $\mathbf{X} = (\mathbf{K},\mathbf{d},{\mathbf{R}_i},{\mathbf{t}_{cp}})$ is the parameter to be refined. $\mathbf{P}_{ij}$ is the $j$-th 3D pattern point of $i$-th image and corresponds to the 2D image point $\mathbf{p}_{ij}$. $\pi(\cdot)$ projects the world points to image according to \cref{eq:KRtP}. $\rho(\cdot)$ is the Cauchy cost function. As is common, we optimize \cref{eq:BA} with the Levenberg-Marquardt (LM) method \cite{LM}, which is implemented in some excellent frameworks\cite{Ceres}.

Note that the camera optical center is fixed at $\mathbf{t}_{cp}$ with the spherical motion constraint. However, for general motion, the camera optical center is free. Therefore, when calibrating the camera using collimator images, the number of motion parameters to be refined is reduced from $6N$ to $3N+3$.

\subsection{Degenerated Configuration}
We prove a degenerate configuration where additional images cannot provide more constraints about the camera intrinsic parameter.

\textbf{Proposition 1.} \textit{If the calibration target rotates only around an axis that is parallel to the $z$ axis and intersects the target plane at $(x,y,0)^T$, the newly added image cannot provide additional constraints on the camera intrinsic parameters.}

Due to space limitations, the proof is provided in the \textit{supplementary material}.
\section{Evaluation and Results}
\label{sec:Experiment}
In this section, we first evaluate the robustness and accuracy of the proposed algorithm on synthetic data. Then, we calibrate the camera and evaluate the effectiveness of the proposed method using real collimator images.
\subsection{Synthetic Data Experiments}
In the synthetic data experiment, we evaluate the performance of the proposed algorithm. We simulate the process of calibrating predefined camera parameters using synthetic images that conform to the spherical motion model. The proposed is compared with the following algorithms: \\
$ \bullet$ \texttt{Zhang}\cite{Zhang2000} is a plane-based algorithm and suitable for general 6DOF motion. \\
$ \bullet$ \texttt{Bouguet}\cite{Bouguet2004} is a calibration toolbox and suitable for general 6DOF motion. Unlike \texttt{Zhang}, \texttt{Bouguet} computes the focal length using the orthogonal vanishing points constraint. \\
$ \bullet$ \texttt{Hartley}\cite{Hartley1997} is a self-calibration method based on a pure rotating camera, which is specifically suitable for the 3DOF spherical motion model.

We define a virtual camera with the following parameters: $f_x = f_y = 1000, c_x = 542, c_y = 478, \gamma = 0.01$, radial distortion coefficients $d_1 = 0.1, d_2 = -0.2$, and the image size is set to $1080 \times 960$ pixels. The calibration target is a planar pattern containing $11 \times 8$ points, each square size of $30 mm \times 30 mm$. The target maintains a spherical motion with a radius of 700$mm$ relative to the camera. The camera optical center in the target coordinate system is $ t_{cp} = [150, 105, -700]^T mm$, and the camera's orientation is arbitrary. Given the camera parameters and poses, the planar target points are projected onto images. Finally, zero-mean Gaussian noise with $\sigma$ standard deviation is added to 2D image points. 
\begin{figure}[tbp]
	\centering
	\includegraphics[width=0.245\linewidth]{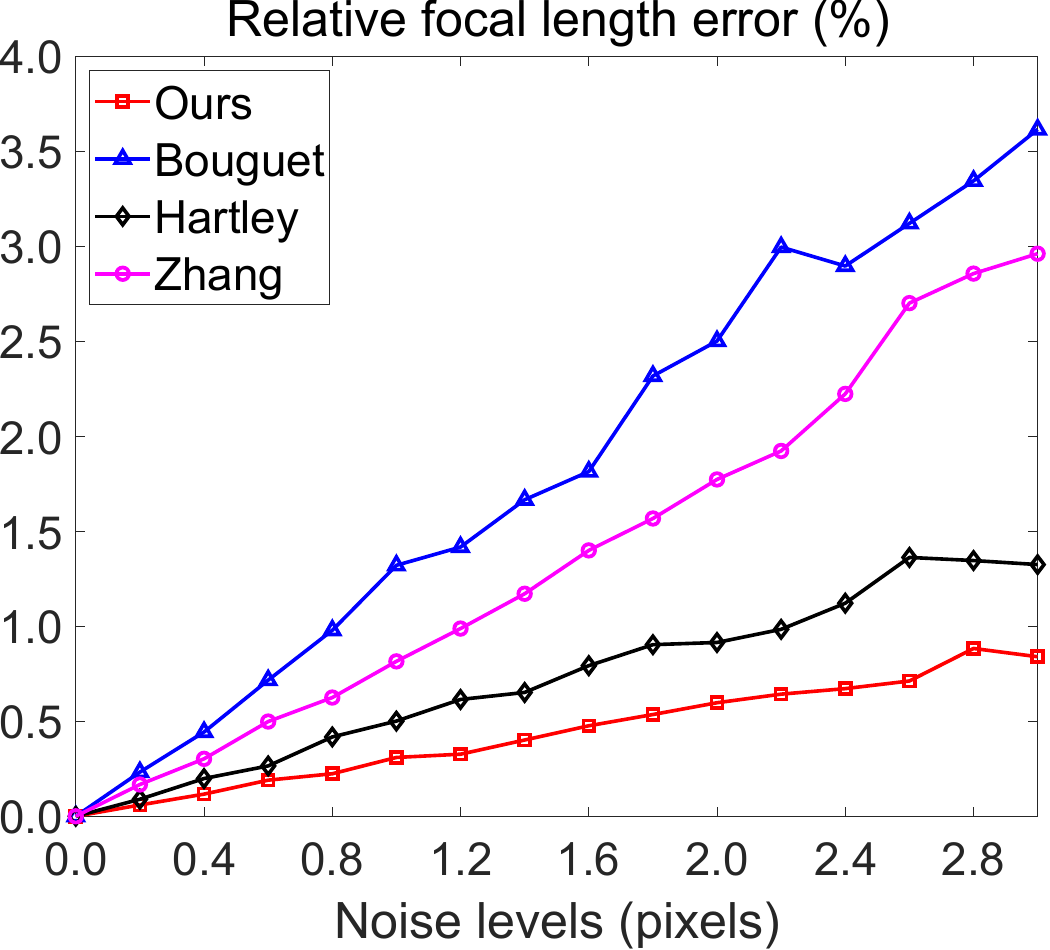} 
	\includegraphics[width=0.235\linewidth]{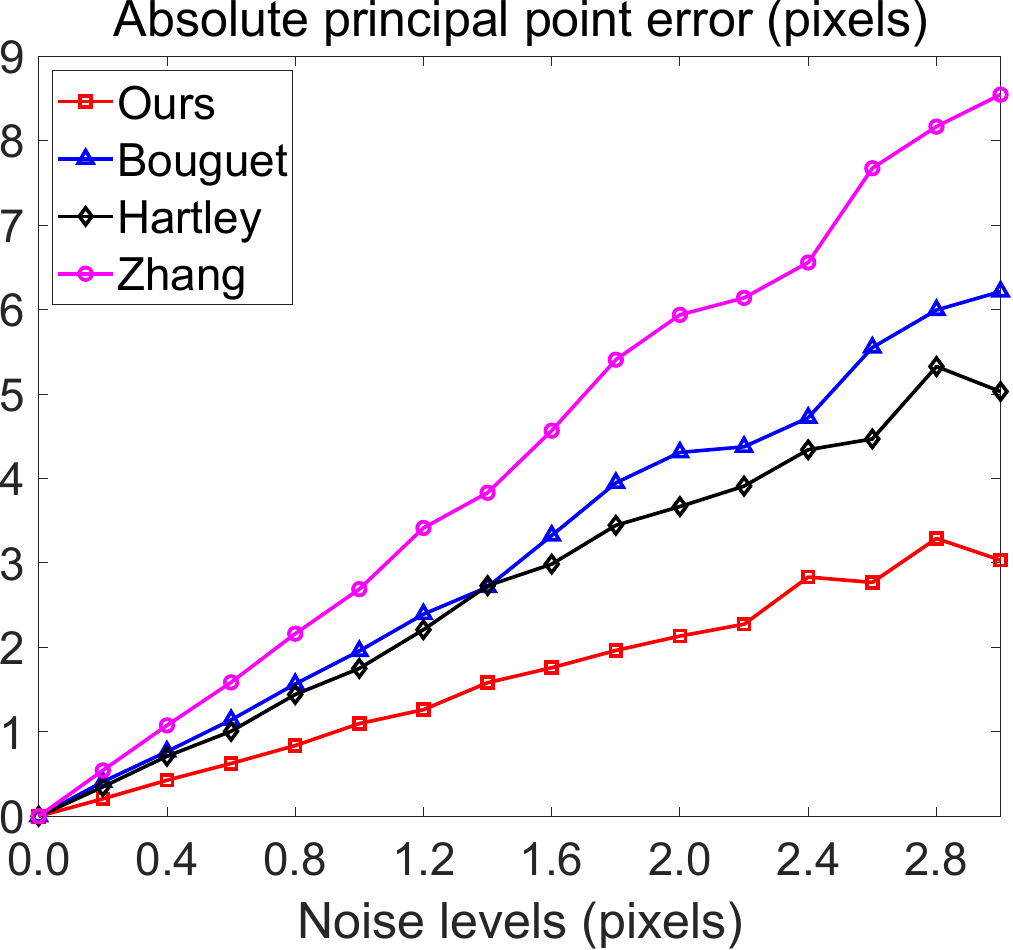} 
	\includegraphics[width=0.245\linewidth]{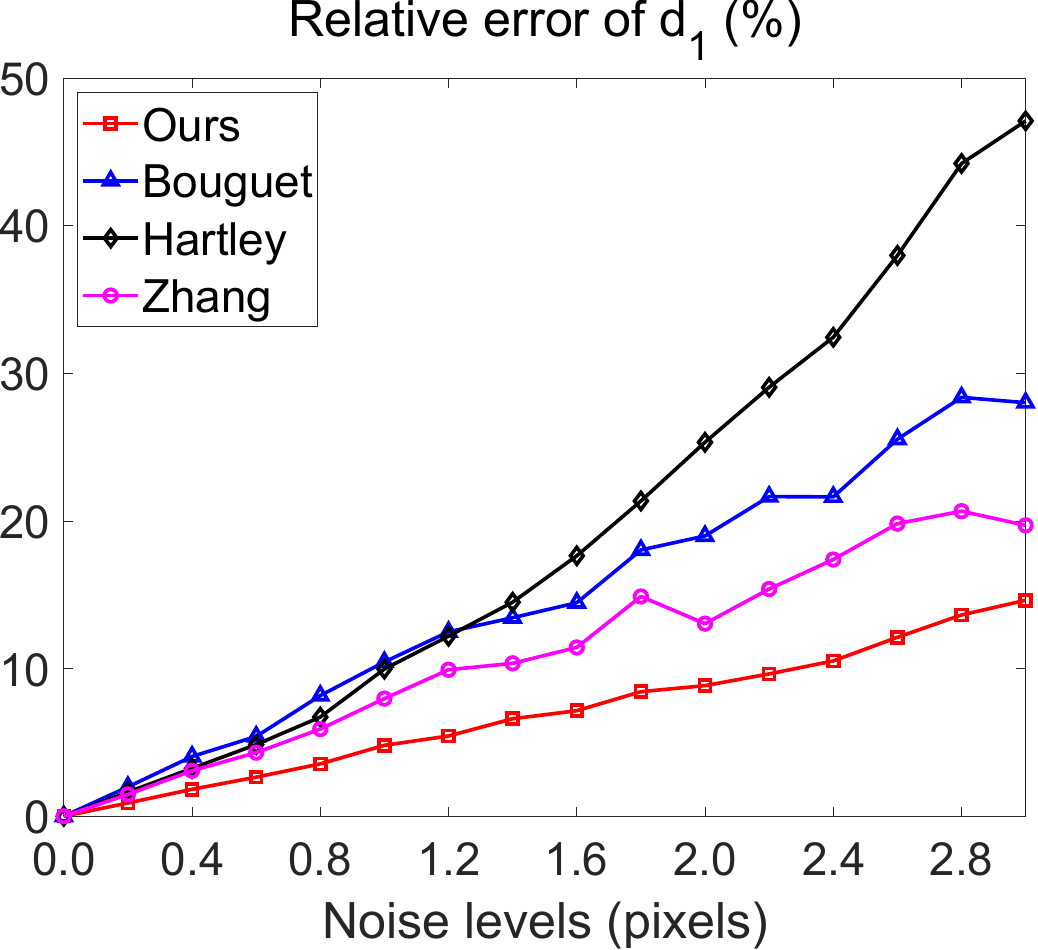}
	\includegraphics[width=0.245\linewidth]{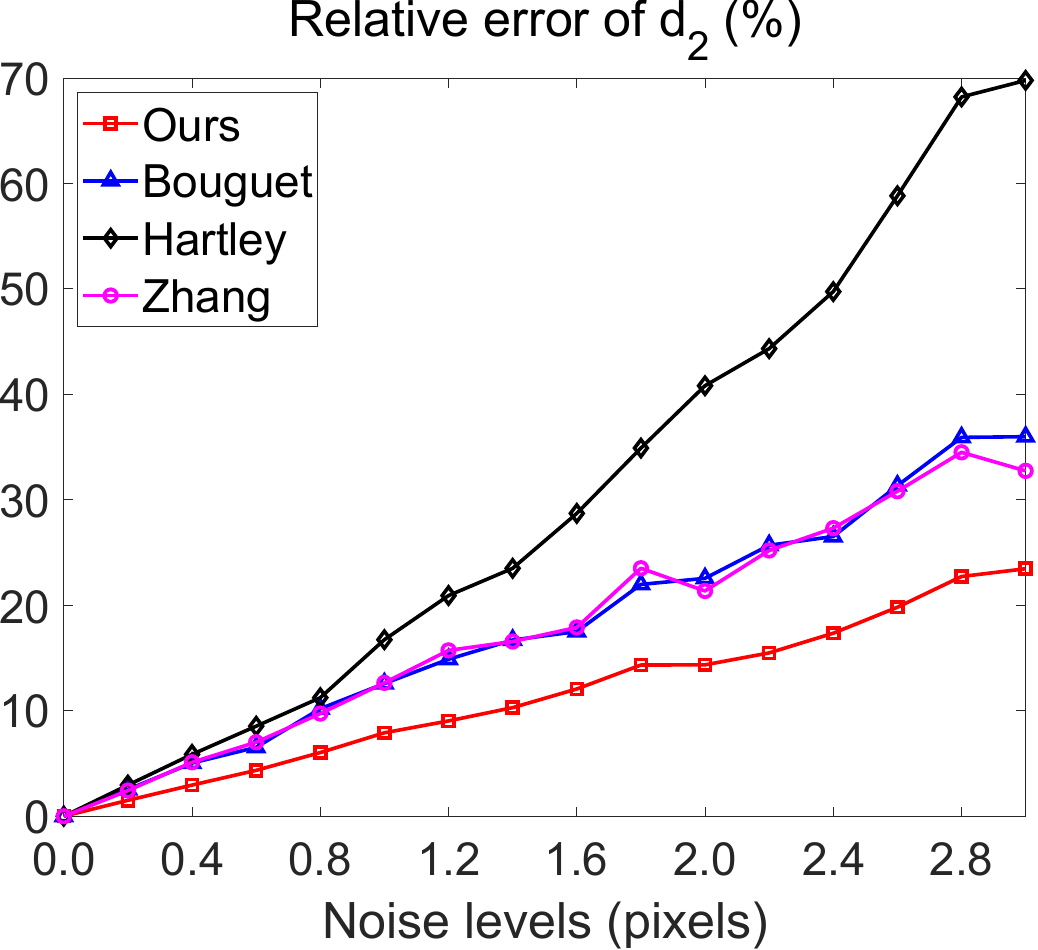}
	\caption{Comparison of calibration errors with increasing noise level.}
	\label{fig:Noise}
\end{figure}
\begin{figure}[tbp]
	\centering
	\includegraphics[width=0.245\linewidth]{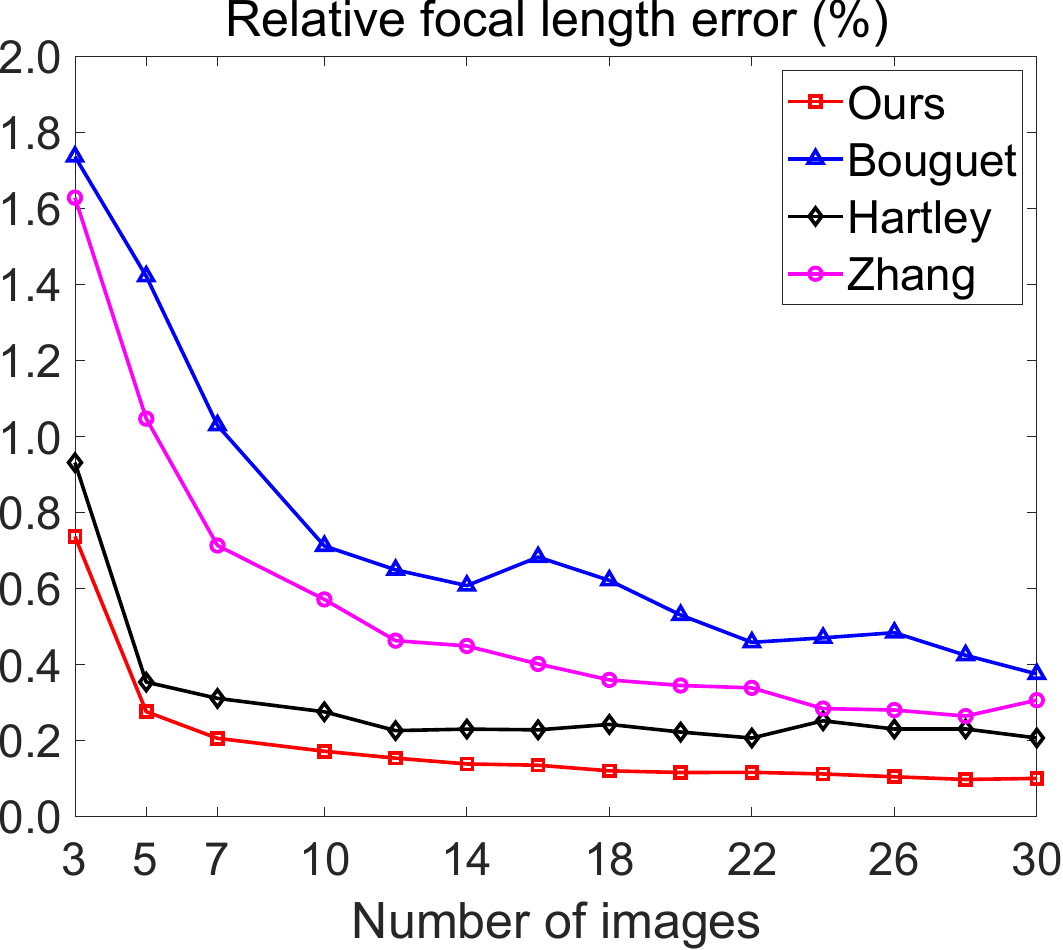}
	\includegraphics[width=0.235\linewidth]{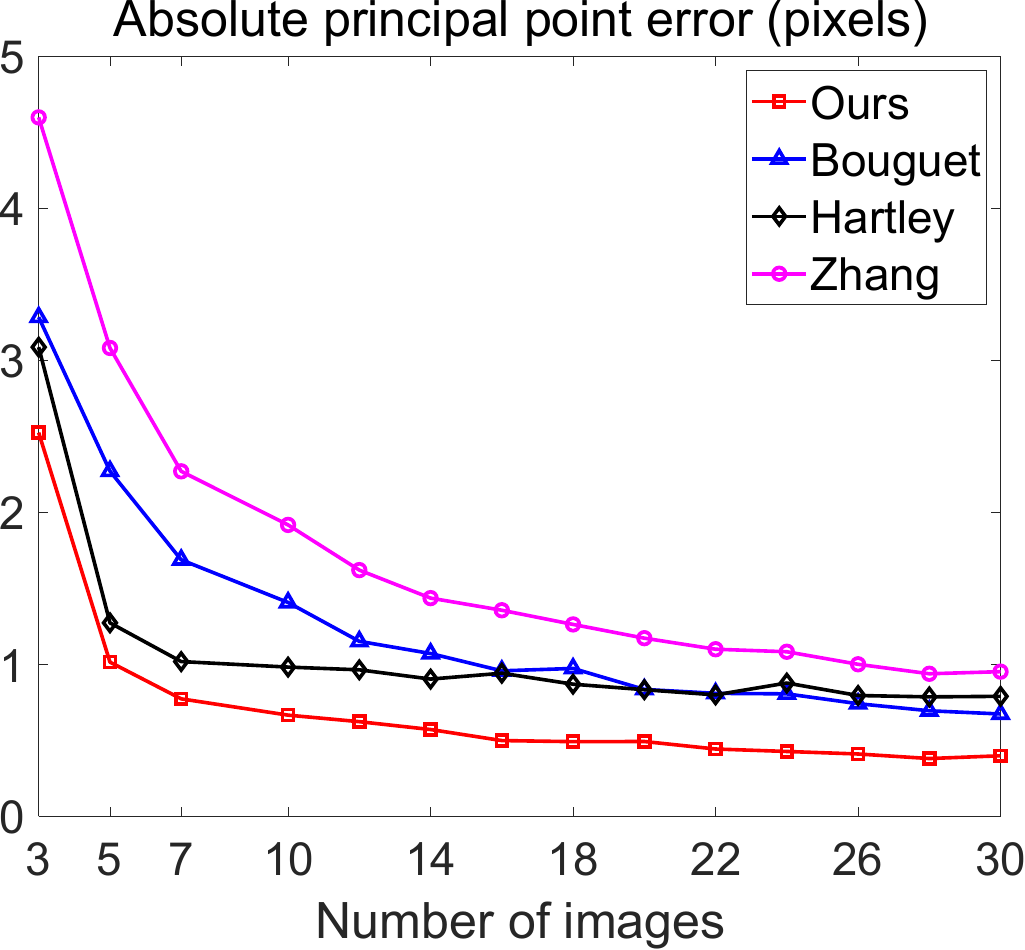} 
	\includegraphics[width=0.245\linewidth]{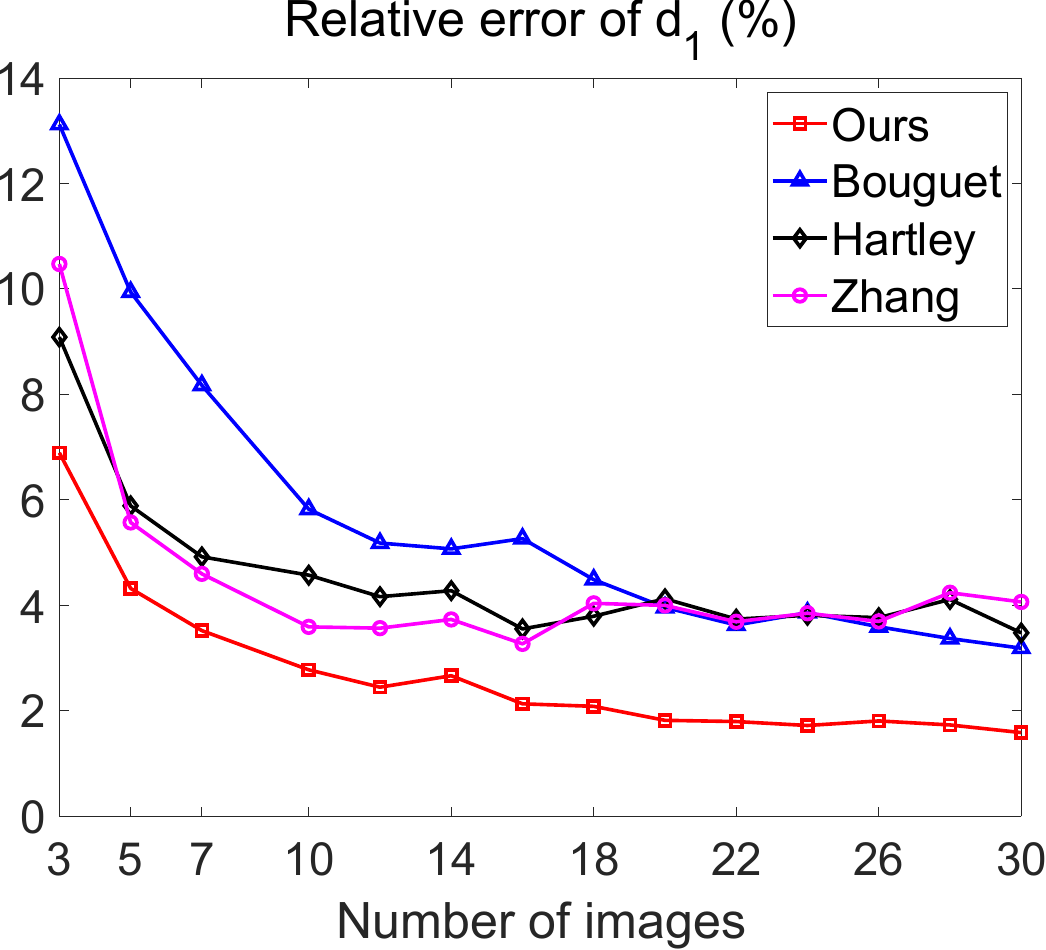}
	\includegraphics[width=0.245\linewidth]{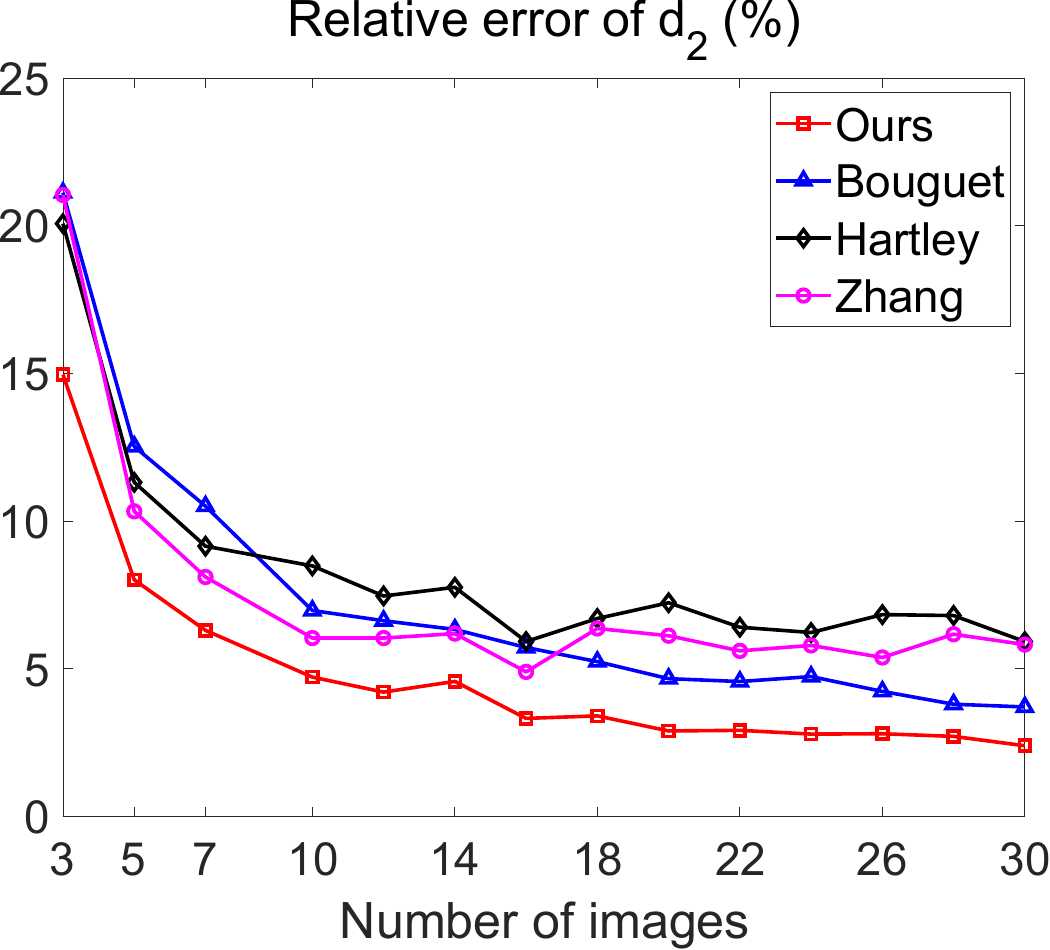}
	\caption{Comparison of calibration errors with the number of images increases.}
	\label{fig:nImgs}
\end{figure}

\textbf{Evaluation of Calibration.} 
In this experiment, we seek to evaluate the performance of the algorithms in solving camera parameters. For the four algorithms, we perform two steps of solving initial value and bundle adjustment to obtain camera parameters. Firstly, we compare the performance of the mentioned algorithms as the noise level increases. The noise level varies from 0 to 3.0 pixels, and we use 15 images to calibrate the camera. For each noise level, we conducted 200 trials and calculated the average errors. As shown in \cref{fig:Noise}, calibration error increases linearly with the noise level. Compared with the other three algorithms, ours shows greater robustness to noise. Additionally, at a noise level of 0.5 pixels, we evaluate the performance with respect to the number of calibration images. The image count ranged from 3 to 30, with 200 trials conducted for each quantity. The results are illustrated in \cref{fig:nImgs}. A significant decrease in error is observed when increasing the number of images from 3 to 5. Both \cref{fig:Noise} and \cref{fig:nImgs} show that our algorithm outperforms other algorithms in robustness and accuracy. We believe that our algorithm benefits from the consideration of both spherical motion and planar homography constraints, along with reducing optimization variables. Due to space limitations, the additional comparison results of the initial value estimation are shown in the \textit{supplementary material}.

\begin{figure}[tbp]
	\centering
	\includegraphics[width=0.3\linewidth]{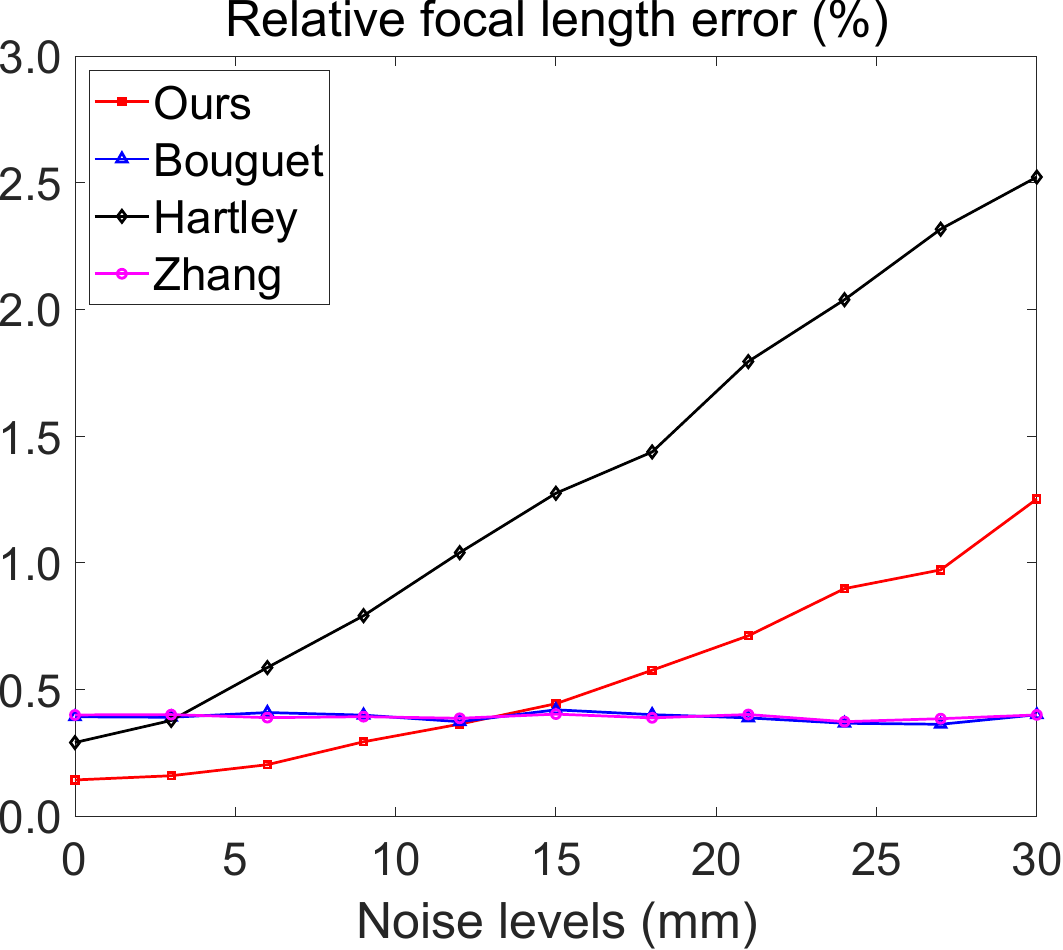} 
	\quad
	\includegraphics[width=0.3\linewidth]{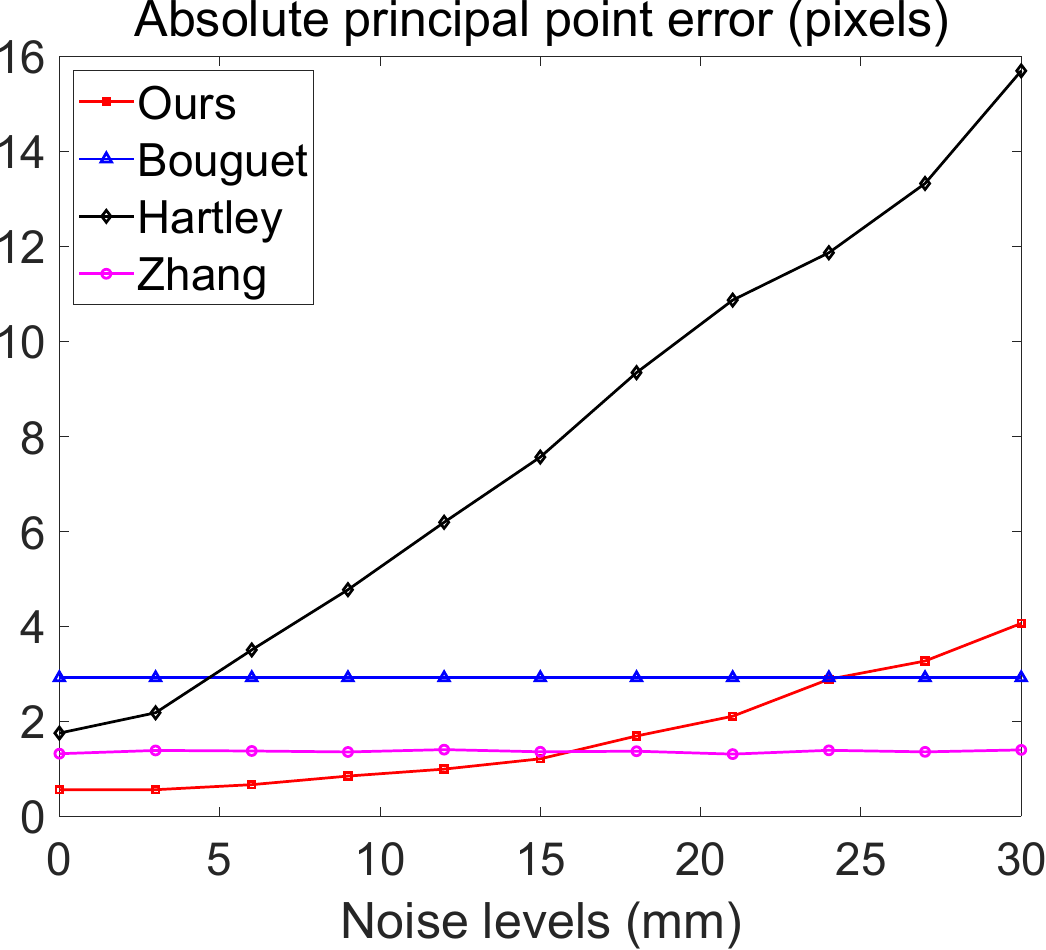}
	\caption{Sensitivity analysis of imperfect spherical motion. }
	\label{fig:Sensi}
\end{figure}

\textbf{Sensitivity of imperfect spherical motion.} 
The low-cost collimators exhibit an inferior parallelism, which is reflected in the fact that the relative motion between the target and camera is not a perfect spherical motion. This experiment evaluates the sensitivity of initial value estimation with respect to imperfect spherical motion. We add zeros-mean Gaussian noise with  standard deviation from $0mm$ to $30mm$ to $\mathbf{t}_{cp}$ to replicate this imperfection. We use 15 images and add a noise level of 0.5 pixels to the image points. For each noise level, a total of independent 500 trials are conducted and the average errors are computed. The sensitivity analysis results are shown in \cref{fig:Sensi}. Noise has no impact on \texttt{Bouguet} and \texttt{Zhang}, because they do not rely on the spherical motion constraint. As the noise increases, calibration errors of \texttt{Ours} and \texttt{Hartley} increase gradually. Notably, \texttt{Hartley} is significantly influenced by noise, whereas \texttt{Ours} displays superior robustness. Even with noise levels below $15mm$ (equivalent to $2.143\%$ of the spherical radius), our algorithm remains superior to the alternatives.

\subsection{Real Images Experiments}
In real image experiments, we verify the effectiveness of our collimator system for calibration and evaluate the performance of the proposed algorithm.

\textbf{Calibration and Accuracy Evaluation.}
Following the experimental methodology utilized in previous studies\cite{Peng2019,Ha2017}, we design a pose estimation experiment to evaluate the accuracy of camera calibration quantitatively. A 8$mm$ camera captures images from our collimator system and a practical printed pattern. Calibration is performed using the collimator images and a subset of the printed pattern images, while the remaining printed pattern images are used to evaluate the calibration results. 
\begin{table}[tbp]
	\centering
	\caption{Comparison of calibration and pose estimation error results (pixels).}
	\resizebox{0.7\linewidth}{!}{
		\begin{tabular}{c|cccccccccccc}
			\hline
			\multirow{2}{*}{Number}& \multicolumn{2}{c}{Zhang\cite{Zhang2000}}&{}&\multicolumn{2}{c}{Bouguet\cite{Bouguet2004}}&{}&\multicolumn{2}{c}{Hartley\cite{Hartley1997}} &{}& \multicolumn{2}{c}{Ours}\\
			\cline { 2 - 3 } \cline { 5 - 6 } \cline { 8 - 9 } \cline { 11 - 12 }
			&{$\epsilon_{calib}$}&{$\epsilon_{eval}$}&{}&{$\epsilon_{calib}$}&{$\epsilon_{eval}$}&{}&{$\epsilon_{calib}$}&{$\epsilon_{eval}$}&{}&{$\epsilon_{calib}$}&{$\epsilon_{eval}$}
			\\
			\hline
			\cellcolor{mypink}{2}&\cellcolor{mypink}{0.1842}&\cellcolor{mypink}{0.1545}&\cellcolor{mypink}{}&\cellcolor{mypink}{0.1739}&\cellcolor{mypink}\textbf{0.1518}&\cellcolor{mypink}{}&\cellcolor{mypink}{-}&\cellcolor{mypink}{-}&\cellcolor{mypink}{}&\cellcolor{mypink}{0.1739}&\cellcolor{mypink}{0.1519}
			\\
			{5}&{0.2316}&{0.1245}&{}&{0.2202}&\textbf{0.1236}&{}&{-}&{0.2577}&{}&{0.2203}&\textbf{0.1236}
			\\
			\cellcolor{mypink}{10}&\cellcolor{mypink}{0.1923}&\cellcolor{mypink}{0.1231}&\cellcolor{mypink}{}&\cellcolor{mypink}{0.1912}&\cellcolor{mypink}{0.1232}&\cellcolor{mypink}{}&\cellcolor{mypink}{-}&\cellcolor{mypink}{0.1986}&\cellcolor{mypink}{}&\cellcolor{mypink}{0.1951}&\cellcolor{mypink}\textbf{0.1222}
			\\
			{15}&{0.1775}&{0.1221}&{}&{0.1718}&{0.1232}&{}&{-}&{0.1791}&{} &{0.1779}&\textbf{0.1216}
			\\
			\cellcolor{mypink}{20}&\cellcolor{mypink}{0.1773}&\cellcolor{mypink}{0.1225}&\cellcolor{mypink}{}&\cellcolor{mypink}{0.1722}&\cellcolor{mypink}{0.1221}&\cellcolor{mypink}{}&\cellcolor{mypink}{-}&\cellcolor{mypink}{0.1871}&\cellcolor{mypink}{} &\cellcolor{mypink}{0.1778}&\cellcolor{mypink}\textbf{0.1207}
			\\
			\hline
		\end{tabular}
	}
	\label{tab:Algo}
\end{table}
After obtaining the camera parameters from the calibration images, the pose of each evaluation image can be determined using absolute pose estimation\cite{OpenGV}.In our setup, a quarter of the corner points on the printed pattern are used for pose estimation, and the remaining points are used for evaluation. By re-projecting the evaluation corner points onto the image with the camera parameters and the calculated pose, we can evaluate the calibration quality based on the re-projection error. A total of 200 printed pattern images are used to evaluate the calibration. The calibration uses 2, 5, 10, 15, and 20 images. The re-projection error of the calibration and evaluation images are donated as $\epsilon_{calib}$ and $\epsilon_{eval}$, respectively. 

The comparison results of the four algorithms are shown in \cref{tab:Algo}. Due to the limited degrees of freedom, our calibrated re-projection error ($\epsilon_{calib}$) is slightly higher. However, \texttt{Ours} exhibits the lowest re-projection error in evaluation images compared to alternative methods, indicating that our method achieves more precise camera parameters. The \texttt{Hartley} method only uses the correspondence of 2D points in images, which makes it impossible to calculate the re-projection error of the calibration images directly. In cases with only two images, \texttt{Zhang} and \texttt{Bouguet} commonly depend on beneficial assumptions, such as $\gamma = 0$. \texttt{Ours} achieves comparable evaluation errors without the need for such assumptions.

Additionally, the \texttt{Bouguet} toolbox is also used to calculate camera parameters from printed pattern images that are not included in the evaluation. We use 20 images for calibration and 200 for evaluation. The estimated camera parameters and error comparison are shown in \cref{tab:Algo2}. The second column is a good reference. The lower evaluation error indicates that our method outperforms the alternative and validates the effectiveness of using the collimator image for calibration. It can be expected that the calibration parameters are not significantly different. Therefore, even a minor improvement holds considerable significance.

Then, we evaluate the performance of all methods using epipolar rectification error, which has not been optimized. Given the camera parameters from different methods, we calculate the epipolar rectification error (in pixels) for 300 pairs of printed evaluation images. The mean and standard deviation of the errors are shown in \cref{tab:ere}. Moreover, the result using \texttt{Bouguet} with 20 printed pattern images is (0.0772, 0.0259) pixels. It is observed that as the number of calibration images increases, the error stabilizes. The results indicate improved accuracy in the collimator image and highlight the beneficial impact of the spherical constraint on accuracy. 
\begin{table}[tbp]
	\centering
	\caption{Comparison of calibration results using 20 images in a real camera.}
	\resizebox{0.65\linewidth}{!}{
		\begin{tabular}{c|c|cccc}
			\hline
			\multirow{2}{*}{}&{Printed}&\multicolumn{3}{c}{Collimator}&{} \\
			\hline
			{}&{Bouguet\cite{Bouguet2004}}&{Zhang\cite{Zhang2000}}&{Bouguet\cite{Bouguet2004}}&{Hartley\cite{Hartley1997}}&{Ours}\\
			\hline
			\cellcolor{mypink}{$f_x$}&\cellcolor{mypink}{2384.5}&\cellcolor{mypink}{2380.8}&\cellcolor{mypink}{2369.2}&\cellcolor{mypink}{2644.6}&\cellcolor{mypink}{2388.3}\\
			{$f_y$}&{2386.3}&{2379.8}&{2368.9}&{2646.0}&{2387.9}\\
			\cellcolor{mypink}{$c_x$}&\cellcolor{mypink}{1209.8}&\cellcolor{mypink}{1222.1}&\cellcolor{mypink}{1221.1}&\cellcolor{mypink}{1214.6}&\cellcolor{mypink}{1221.0}\\
			{$c_y$}&{1008.2}&{1008.4}&{1009.8}&{1012.7}&{1009.6}\\
			\cellcolor{mypink}{$k_1$}&\cellcolor{mypink}{-0.0843}&\cellcolor{mypink}{-0.0899}&\cellcolor{mypink}{-0.0908}&\cellcolor{mypink}{-0.0754}&\cellcolor{mypink}{-0.0900}\\
			{$k_2$}&{0.0819}&{0.0912}&{0.0892}&{0.1181}&{0.0954}\\
			\hline
			\cellcolor{mypink}{$\epsilon_{calib}$}&\cellcolor{mypink}{0.1513}&\cellcolor{mypink}{0.1773}&\cellcolor{mypink}{0.1722}&\cellcolor{mypink}{-}&\cellcolor{mypink}{0.1778}\\
			{$\epsilon_{eval}$}&{0.1225}&{0.1225}&{0.1221}&{0.1871}&{0.1207}\\
			\hline
		\end{tabular}
	}
	\label{tab:Algo2}
\end{table}
\begin{table}[tbp]
	\centering
	\caption{Comparison of epipolar rectification error results (pixels).}
	\resizebox{0.7\linewidth}{!}{
		\begin{tabular}{c|cccccccccccc}
			\hline
			\multirow{2}{*}{Number}& \multicolumn{2}{c}{Zhang\cite{Zhang2000}}&{}&\multicolumn{2}{c}{Bouguet\cite{Bouguet2004}}&{}&\multicolumn{2}{c}{Hartley\cite{Hartley1997}} &{}& \multicolumn{2}{c}{Ours}\\
			\cline { 2 - 3 } \cline { 5 - 6 } \cline { 8 - 9 } \cline { 11 - 12 }
			&{mean}&{std}&{}&{mean}&{std}&{}&{mean}&{std}&{}&{mean}&{std}
			\\
			\hline
			\cellcolor{mypink}{2}&\cellcolor{mypink}{0.0846}&\cellcolor{mypink}{0.0406}&\cellcolor{mypink}{}&\cellcolor{mypink}\textbf{0.0770}&\cellcolor{mypink}{0.0220}&\cellcolor{mypink}{}&\cellcolor{mypink}{-}&\cellcolor{mypink}{-}&\cellcolor{mypink}{}&\cellcolor{mypink}\textbf{0.0770}&\cellcolor{mypink}{0.0229}
			\\
			{5}&{0.0809}&{0.0307}&{}&{0.0764}&{0.0214}&{}&{0.0832}&{0.0397}&{}&\textbf{0.0762}&{0.0211}
			\\
			\cellcolor{mypink}{10}&\cellcolor{mypink}{0.0773}&\cellcolor{mypink}{0.0234}&\cellcolor{mypink}{}&\cellcolor{mypink}{0.0758}&\cellcolor{mypink}{0.0213}&\cellcolor{mypink}{}&\cellcolor{mypink}{0.0854}&\cellcolor{mypink}{0.0509}&\cellcolor{mypink}{}&\cellcolor{mypink}\textbf{0.0757}&\cellcolor{mypink}{0.0208}
			\\
			{15}&{0.0782}&{0.0266}&{}&{0.0771}&{0.0253}&{}&{0.0837}&{0.0522}&{} &\textbf{0.0756}&{0.0207}
			\\
			\cellcolor{mypink}{20}&\cellcolor{mypink}{0.0772}&\cellcolor{mypink}{0.0252}&\cellcolor{mypink}{}&\cellcolor{mypink}{0.0768}&\cellcolor{mypink}{0.0253}&\cellcolor{mypink}{}&\cellcolor{mypink}{0.0818}&\cellcolor{mypink}{0.0365}&\cellcolor{mypink}{} &\cellcolor{mypink}\textbf{0.0756}&\cellcolor{mypink}{0.0208}
			\\
			\hline
		\end{tabular}}
	\label{tab:ere}
\end{table}
\begin{figure}[tbp]
	\centering
	\begin{subfigure}[t]{\linewidth}
		\centering
		\includegraphics[width=0.5\linewidth]{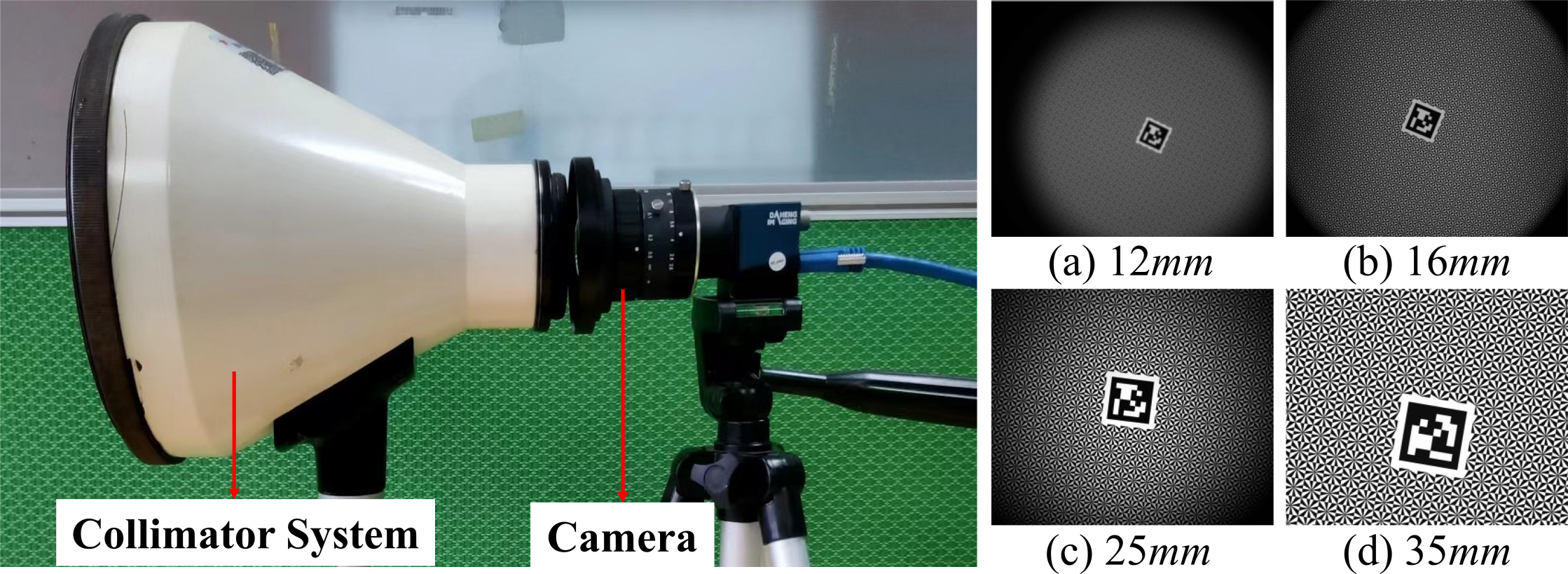}
	\end{subfigure}%
	\caption{ Calibration image sample obtained from the collimator.}
	\label{fig:SampleImages}
\end{figure}
\begin{table}[tbp]
	\centering
	\caption{Comparison of calibration results with different focal length lense (pixels).} 	
	\begin{tabular}{c|cccc}
		\hline
		{}&{Zhang\cite{Zhang2000}}&{Bouguet\cite{Bouguet2004}}&{Hartley\cite{Hartley1997}}&{Ours}\\
		\hline
		\cellcolor{mypink}{12$mm$}&\cellcolor{mypink}{0.2273}&\cellcolor{mypink}{0.2225}&\cellcolor{mypink}{0.3404}&\cellcolor{mypink}\textbf{0.2174}
		\\
		{16$mm$}&{0.2994}&{0.2591}&{0.4182}&\textbf{0.2536}
		\\
		\cellcolor{mypink}{25$mm$}&\cellcolor{mypink}{0.3133}&\cellcolor{mypink}{0.3078}&\cellcolor{mypink}{0.4195}&\cellcolor{mypink}\textbf{0.2923}
		\\
		{35$mm$}&{0.3289}&{0.3348}&{0.7272}&\textbf{0.3118}
		\\
		\hline
	\end{tabular}
	\label{tab:Lenses}
\end{table}

\textbf{ Multiple Cameras Calibration with a Same Collimator.} 
In this experiment, we evaluate our method on multiple cameras with the same collimator. We used the cameras with 12$mm$, 16$mm$, 25$mm$, and 35$mm$ focal lengths of lenses. The calibration images are captured from the same collimator, as shown in \cref{fig:SampleImages} and left of \cref{fig:coll_calib}. As the focal length increases, the more pixels a single square occupies. Cameras with different focal lengths can observe the calibration target through our collimator system in a limited space. 

Similar to the evaluation approach in the previous experiment, we segregate all collimator images into calibration and evaluation sets. For each lens, we use 20 images for calibration and 200 for evaluation. The results are listed in \cref{tab:Lenses}. The proposed algorithm demonstrates the lowest re-projection error for lenses with varying focal lengths among the evaluation images. The re-projection error tends to increase with the increase of focal length. We believe that the main reason is that the longer the lens's focal length, the fewer points observed. More importantly, this experiment confirms the effectiveness of our collimator system and the proposed method. Moreover, we provide a spherical motion verification experiment in \textit{supplementary material}.

\section{Conclusion}
This paper presents a novel camera calibration method using a designed collimator system. Based on the optical geometry of the collimator system, we prove that the relative motion between the target and the camera conforms to a spherical motion. Furthermore, we propose a closed-form solver for multiple views and a minimal solver for two views for camera calibration. Our collimator-based method effectively combines the advantages of close-range calibration using a collimator and flexibility using a planar calibration target. We validate the effectiveness of using only collimator images to calibrate the camera. Synthetic and real data experiments demonstrate that the calibration accuracy of the proposed method is superior to the state-of-the-art methods.

\textbf{Limitations:} When calibrating a fisheye camera, the distortion should be paid more attention rather than simply setting it to $\mathbf{0}$. Future work will focus on cameras with significant distortion.

\section*{Acknowledgements}
This research has been supported by the Hunan Provincial Natural Science Foundation for Excellent Young Scholars under Grant 2023JJ20045, and the National Natural Science Foundation of China under Grant 12372189. We thank Zijie Wang for useful discussion on the writing of the paper.

{
	\bibliographystyle{splncs04}
	\bibliography{main}

\begin{thebibliography}{10}
\providecommand{\url}[1]{\texttt{#1}}
\providecommand{\urlprefix}{URL }
\providecommand{\doi}[1]{https://doi.org/#1}

\bibitem{Ceres}
Agarwal, S., Mierle, K., Team, T.C.S.: {Ceres Solver}.
  \url{https://github.com/ceres-solver/ceres-solver}

\bibitem{Bauer2008}
Bauer, M., Grießbach, D., Hermerschmidt, A., Krüger, S., Scheele, M.,
  Schischmanow, A.: Geometrical camera calibration with diffractive optical
  elements. Optics express  \textbf{16}(25),  20241--20248 (2008)

\bibitem{Bouguet2004}
Bouguet, J.Y.: Camera calibration toolbox for matlab.
  \url{http://robots.stanford.edu/cs223b04/JeanYvesCalib/index.html} (2004)

\bibitem{Opencv2000}
Bradski, G.: The opencv library. Dr. Dobb's Journal: Software Tools for the
  Professional Programmer  \textbf{25}(11),  120--123 (2000)

\bibitem{Brown1971}
Brown, D.C.: Close-range camera calibration. Photogramm. Eng  \textbf{37}(8),
  855--866 (1971)

\bibitem{Campos2021}
Campos, C., Elvira, R., Rodríguez, J.J.G., M.~Montiel, J.M., D.~Tardós, J.:
  Orb-slam3: An accurate open-source library for visual, visual–inertial, and
  multimap slam. IEEE Transactions on Robotics  \textbf{37}(6),  1874--1890
  (2021)

\bibitem{Clarke1998}
Clarke, T.A., Fryer, J.G.: The development of camera calibration methods and
  models. The Photogrammetric Record  \textbf{16}(91),  51--66 (1998)

\bibitem{Berlin1992}
Faugeras, O.D., Luong, Q.T., Maybank, S.J.: Camera self-calibration: Theory and
  experiments. In: IEEE European Conference on Computer Vision (ECCV). p.
  321–334 (1992)

\bibitem{Gustavo2017}
Führ, G., Jung, C.R.: Camera self-calibration based on nonlinear optimization
  and applications in surveillance systems. IEEE Transactions on Circuits and
  Systems for Video Technology  \textbf{27}(5),  1132--1142 (2017)

\bibitem{Guan2022}
Guan, B., Su, Z., Yu, Q., Li, Z., Feng, W., Yang, D., Zhang, D.: Monitoring the
  blades of a wind turbine by using videogrammetry. Optics and Lasers in
  Engineering  \textbf{152},  106901 (2022)

\bibitem{Guan2023}
Guan, B., Zhao, J., Barath, D., Fraundorfer, F.: Minimal solvers for relative
  pose estimation of multi-camera systems using affine correspondences.
  International Journal of Computer Vision  \textbf{131}(1),  324--345 (2023)

\bibitem{GuanTCYB2021}
Guan, B., Zhao, J., Li, Z., Sun, F., Fraundorfer, F.: Relative pose estimation
  with a single affine correspondence. IEEE Transactions on Cybernetics
  \textbf{52}(10),  10111--10122 (2022)

\bibitem{Ha2017}
Ha, H., Perdoch, M., Alismail, H., Kweon, I.S., Sheikh, Y.: Deltille grids for
  geometric camera calibration. In: IEEE International Conference on Computer
  Vision (ICCV). pp. 5354--5362 (2017)

\bibitem{Hartley2012}
Hartley, R., Li, H.: An efficient hidden variable approach to minimal-case
  camera motion estimation. IEEE Transactions on Pattern Analysis and Machine
  Intelligence  \textbf{34}(12),  2303--2314 (2012)

\bibitem{Hartley2004}
Hartley, R., Zisserman, A.: Multiple View Geometry in Computer Vision.
  Cambridge University Press, 2 edn. (2004)

\bibitem{Hartley1994}
Hartley, R.I.: Self-calibration from multiple views with a rotating camera. In:
  European Conference on Computer Vision (ECCV). pp. 471--478 (1994)

\bibitem{Hartley1997}
Hartley, R.I.: Self-calibration of stationary cameras. International Journal of
  Computer Vision  \textbf{22}(1),  5--23 (1997)

\bibitem{Heikkila2000}
Heikkila, J.: Geometric camera calibration using circular control points. IEEE
  Transactions on Pattern Analysis and Machine Intelligence  \textbf{22}(10),
  1066--1077 (2000)

\bibitem{Herrera2016}
Herrera, D., Kannala, C.J., Heikkila, J.: Forget the checkerboard: Practical
  self-calibration using a planar scene. In: IEEE Winter Conference on
  Applications of Computer Vision (WACV). pp.~1--9. IEEE (2016)

\bibitem{Hieronymus2012}
Hieronymus, J.: Comparison of methods for geometric camera calibration. The
  International Archives of the Photogrammetry, Remote Sensing and Spatial
  Information Sciences  \textbf{39},  595--599 (2012)

\bibitem{Hu2022}
Hu, H., Wei, B., Mei, S., Liang, J., Zhang, Y.: A two-step calibration method
  for vision measurement with large field of view. IEEE Transactions on
  Instrumentation and Measurement  \textbf{71},  1--10 (2022)

\bibitem{Kannala2006}
Kannala, J., Brandt, S.S.: A generic camera model and calibration method for
  conventional, wide-angle, and fish-eye lenses. IEEE Transactions on Pattern
  Analysis and Machine Intelligence  \textbf{28}(8),  1335--1340 (2006)

\bibitem{Klaus2004}
Klaus, A., Bauer, J., Karner, K., Elbischger, P., Perko, R., Bischof, H.:
  Camera calibration from a single night sky image. In: IEEE Conference on
  Computer Vision and Pattern Recognition (CVPR). vol.~1, pp.~I--I (2004)

\bibitem{OpenGV}
Kneip, L., Furgale, P.: Opengv: A unified and generalized approach to real-time
  calibrated geometric vision. In: IEEE International Conference on Robotics
  and Automation (ICRA). pp.~1--8 (2014)

\bibitem{Liu2015}
Liu, Z., Li, F., Li, X., Zhang, G.: A novel and accurate calibration method for
  cameras with large field of view using combined small targets. Measurement
  \textbf{64},  1--16 (2015)

\bibitem{Lochman2021}
Lochman, Y., Liepieshov, K., Chen, J., Perdoch, M., Zach, C., Pritts, J.:
  Babelcalib: A universal approach to calibrating central cameras. In: IEEE
  International Conference on Computer Vision (ICCV). pp. 15253--15262 (2021)

\bibitem{LM}
Mor{\'e}, J.J.: The levenberg-marquardt algorithm: implementation and theory.
  In: Numerical analysis: proceedings of the biennial Conference held at
  Dundee, June 28--July 1, 1977. pp. 105--116 (2006)

\bibitem{Nister2005}
Nist{\'e}r, D., Stew{\'e}nius, H., Grossmann, E.: Non-parametric
  self-calibration. In: IEEE International Conference on Computer Vision
  (ICCV). vol.~1, pp. 120--127 (2005)

\bibitem{Oniga2018}
Oniga, V.E., Pfeifer, N., Loghin, A.M.: 3d calibration test-field for digital
  cameras mounted on unmanned aerial systems (uas). Remote Sensing
  \textbf{10}(12), ~2017 (2018)

\bibitem{Peng2019}
Peng, S., Sturm, P.: Calibration wizard: A guidance system for camera
  calibration based on modelling geometric and corner uncertainty. In: IEEE
  International Conference on Computer Vision (ICCV). pp. 1497--1505 (2019)

\bibitem{Pollefeys1999}
Pollefeys, M., Koch, R., Gool, L.V.: Self-calibration and metric reconstruction
  inspite of varying and unknown intrinsic camera parameters. International
  journal of computer vision  \textbf{32}(1),  7--25 (1999)

\bibitem{Colmap2016}
Schönberger, J.L., Frahm, J.M.: Structure-from-motion revisited. In: IEEE
  Conference on Computer Vision and Pattern Recognition (CVPR). pp. 4104--4113
  (2016)

\bibitem{Thomas2020}
Schöps, T., Larsson, V., Pollefeys, M., Sattler, T.: Why having 10,000
  parameters in your camera model is better than twelve. In: IEEE Conference on
  Computer Vision and Pattern Recognition (CVPR). pp. 2532--2541 (2020)

\bibitem{Shang2013}
Shang, Y., Sun, X., Yang, X., Wang, X., Yu, Q.: A camera calibration method for
  large field optical measurement. Optik - International Journal for Light and
  Electron Optics  \textbf{124},  6553--6558 (12 2013)

\bibitem{Sturm1999}
Sturm, P.F., Maybank, S.J.: On plane-based camera calibration: A general
  algorithm, singularities, applications. In: IEEE Conference on Computer
  Vision and Pattern Recognition (CVPR). vol.~1, pp. 432--437. IEEE (1999)

\bibitem{Triggs2000}
Triggs, B., McLauchlan, P.F., Hartley, R.I., Fitzgibbon, A.W.: Bundle
  adjustment—a modern synthesis. In: Vision Algorithms: Theory and Practice:
  International Workshop on Vision Algorithms. pp. 298--372. Springer (2000)

\bibitem{Tsai1987}
Tsai, R.: A versatile camera calibration technique for high-accuracy 3d machine
  vision metrology using off-the-shelf tv cameras and lenses. IEEE Journal on
  Robotics and Automation  \textbf{3}(4),  323--344 (1987)

\bibitem{Ventura2016}
Ventura, J.: Structure from motion on a sphere. In: European Conference on
  Computer Vision (ECCV). vol.~9907, pp. 53--68 (2016)

\bibitem{Wang2022}
Wang, C., Liu, Y., Wang, Y., Li, X., Wang, M.: Efficient and outlier-robust
  simultaneous pose and correspondence determination by branch-and-bound and
  transformation decomposition. IEEE Transactions on Pattern Analysis and
  Machine Intelligence  \textbf{44}(10),  6924--6938 (2022)

\bibitem{Wang2015}
Wang, Z., Wu, Z., Zhen, X., Yang, R., Xi, J., Chen, X.: A two-step calibration
  method of a large fov binocular stereovision sensor for onsite measurement.
  Measurement  \textbf{62},  15--24 (2015)

\bibitem{Wu2007}
Wu, G., Han, B., He, X.: Calibration of geometric parameters of line array ccd
  camera based on exact measuring angle in lab. Optics and Precision
  Engineering  \textbf{15}(10),  1628--1632 (2007)

\bibitem{Xiao2010}
Xiao, Z., Jin, L., Yu, D., Tang, Z.: A cross-target-based accurate calibration
  method of binocular stereo systems with large-scale field-of-view.
  Measurement  \textbf{43}(6),  747--754 (2010)

\bibitem{Yuan2021}
Yuan, G., Zheng, L., Ding, Y., Zhang, H., Zhang, X., Liu, X., Sun, J.: A
  precise calibration method for line scan cameras. IEEE Transactions on
  Instrumentation and Measurement  \textbf{70}, ~1--9 (2021)

\bibitem{Yuan2019}
Yuan, G., Zheng, L., Sun, J., Liu, X., Wang, X., Zhang, Z.: Practical
  calibration method for aerial mapping camera based on multiple pinhole
  collimator. IEEE Access  \textbf{8},  39725--39733 (2019)

\bibitem{Zhang2020}
Zhang, K., Xie, J., Snavely, N., Chen, Q.: Depth sensing beyond lidar range.
  In: IEEE Conference on Computer Vision and Pattern Recognition (CVPR). pp.
  1689--1697 (2020)

\bibitem{Zhang2000}
Zhang, Z.: A flexible new technique for camera calibration. IEEE Transactions
  on Pattern Analysis and Machine Intelligence  \textbf{22}(11),  1330--1334
  (2000)

\bibitem{Zhang2004}
Zhang, Z.: Camera calibration with one-dimensional objects. IEEE Transactions
  on Pattern Analysis and Machine Intelligence  \textbf{26}(7),  892--899
  (2004)

\end{thebibliography}
}
\clearpage
\title{Camera Calibration using a Collimator System \\
	\vspace{10pt} 
	 \large{Supplementary Materials}}
\author{}
\institute{}
\maketitle

In the supplementary material, we present the proof of the degenerate configuration (see \cref{sec:PSM}) and the proof of spherical motion (see \cref{sec:PDC}). The performance of initial value estimation is evaluated in \cref{sec:PIVE}. \cref{sec:VSM} verifies the spherical motion model using real collimator images. Results of the structure from motion are shown in \cref{sec:SfM}

\section{Proof of Spherical Motion}
\label{sec:PSM}
In Section 3.2 of the paper, we propose that the relative motion between the calibration target and the camera conforms to the spherical motion model. This section gives more details of the proof. By arbitrarily selecting a pair of planar points $\mathbf{P}_i$ and $\mathbf{P}_j$ on the target and transforming them, we can have the following equation from the angle invariance:
\begin{equation}
	\angle\left(\mathbf{P}_i , \mathbf{P}_j \right) = \angle\left(\mathbf{R}\left(\mathbf{P}_i+\mathbf{t}\right), \mathbf{R}\left(\mathbf{P}_j +\mathbf{t}\right)\right) 
	\label{eq:RPt1}
\end{equation}
where $\mathbf{R}$ is rotation matrix and $\mathbf{t}$ is translation vector. According to the rotation invariance\cite{Wang2022}, a pair of 3D points are jointly rotated about the origin, their angular distance remains unchanged. The angular distance means the angle between two vectors from the origin to two points. Therefore, we have:
\begin{equation}
\angle(\mathbf{P}_i, \mathbf{P}_j)) = \angle(\mathbf{P}_i+\mathbf{t}, \mathbf{P}_j +\mathbf{t}).
\label{eq:RPt2}
\end{equation}
To make \cref{eq:RPt2} hold for any pair of points, the translation must be $\mathbf{0}_{3\times1}$ vector ($\mathbf{t}=[0,0,0]^T$). We select several special point pairs to support our proof. Our proof is similar to the proof by contradiction. Angle invariance holds for any point pair on the target, so it must hold for specific pairs. We prove that for specific point pairs, the condition for angle invariance is $\mathbf{t} = \mathbf{0}$. If $\mathbf{t}$ is non-zero, angle invariance is not maintained. The detailed proof is given as follows.

Without loss of generality, we establish a coordinate system with the camera optical center as the origin. The $z$-axis is perpendicular to the plane of the target, and the $xy$-plane is parallel to the target. We assume that the distance from the origin to the target is $r(r>0)$. 

Since \cref{eq:RPt2} holds for any pair of points, we choose two special pairs of points on the target:
\begin{align*}
	\mathbf{A} &= (r,0,r)^T, \quad \ \, \mathbf{B} = (-r,0,r)^T, \\
	\mathbf{C} &= (2r,0,r)^T,\quad \mathbf{D} = (-2r,0,r)^T. \\
\end{align*}
The angle distance of $(\mathbf{A},\mathbf{B})$ and $(\mathbf{C},\mathbf{D})$ can be easily calculated:
\begin{equation}
	\cos(\angle(\mathbf{A},\mathbf{B})) = 0, \quad \cos(\angle(\mathbf{C},\mathbf{D})) = -\frac{3}{5}.
\end{equation}
Let us define a $\mathbf{t}=(x,y,z)^T$ to represent the relative translation between the target and the origin. From \cref{eq:RPt2}, we have:
\begin{equation}
\cos(\angle(\mathbf{A},\mathbf{B})) = \cos(\angle(\mathbf{A}+\mathbf{t},\mathbf{B}+\mathbf{t}))= x^2 - r^2 +y^2+(r+z)^2 = 0, \label{eq:AtBt}
\end{equation}

\begin{equation}
	\begin{aligned}
		\quad \	\cos(\angle(\mathbf{C},\mathbf{D})) &= \cos(\angle(\mathbf{C}+\mathbf{t},\mathbf{D}+\mathbf{t})) \\ 
		&= \frac{x^2 - 4r^2 +y^2+(r+z)^2 }{\sqrt{(x+2r)^2 +y^2+(r+z)^2}\sqrt{(x-2r)^2 +y^2+(r+z)^2}} \\
		& =  -\frac{3}{5}.
	\end{aligned}
\label{eq:CtDt}
\end{equation}
Substituting \cref{eq:AtBt} into \cref{eq:CtDt}, we get:
\begin{align}
	\frac{-3r^2}{\sqrt{25r^4-16x^2r^2}} &= -\frac{3}{5} \\
	\Rightarrow \frac{9r^4}{25r^4-16x^2r^2} &= \frac{9}{25} \\
	\Rightarrow \frac{9}{25-16\frac{x^2}{r^2}} &= \frac{9}{25}.
\end{align}
Obviously, we get $\frac{x^2}{r^2}=0$ and $x=0$. Similarly, we can also calculate $y=0$ when we select the following pairs of points:
\begin{align*}
	\mathbf{E} &= (0,r,r)^T, \quad \ \, \mathbf{F} = (0,-r,r)^T, \\
	\mathbf{G} &= (0,2r,r)^T,\quad \mathbf{H} = (0,-2r,r)^T. \\
\end{align*}

Then, substituting $x=0$ and $y=0$ into \cref{eq:AtBt}, we have $(r+z)^2 = r^2$ and then get $r=0$. Therefore, to make \cref{eq:RPt2} holds for all pairs on the target, the translation vector $\mathbf{t}$ is equal to $\mathbf{0}$ ($\mathbf{t}=[0,0,0]^T$). Up to now, we have proven that the relative motion between the target and camera optical center is a pure rotation motion without relative translation, called spherical motion. 

\section{Proof of Degenerate Configuration}
\label{sec:PDC}

In Section 4.5 of the paper, we propose a degenerate configuration of camera calibration with spherical motion constraints. This section gives more details and proof of the degenerate configuration. Inspired by \cite{Zhang2000}, we believe the newly added image is degraded when the image cannot provide more constraints for solving the camera parameters. In the spherical motion model, the calibration target exhibits only 3DOF rotational motion with respect to the camera. Therefore, we can prove a degenerate configuration in the following.

\textbf{Proposition 1.} \textit{If the calibration target rotates only around an axis that is parallel to the $z$ axis and intersects the target plane at $(x,y,0)^T$, the newly added image cannot provide additional constraints on the camera intrinsic parameters.}

\textit{Proof.} We will use the superscript ($^\prime$) to distinguish the matrix related to the newly added and original images. According to Eq. (4) in the paper, we suppose that the pose of the target in the original image is represented as follows:
\begin{equation}
	\mathbf{T}_{pc} = \left[ {\begin{array}{*{20}{c}}
			{{\mathbf{R}_{pc}}}&{ - {\mathbf{R}_{pc}}{\mathbf{t}_{cp}}}\\
			{\mathbf{0}_{1 \times 3}}&1
	\end{array}} \right],
	\label{T_pc}
\end{equation}
where ${\mathbf{t}_{cp}} = (x,y,r)^T$ and $\mathbf{R}_{pc}$ is a $3 \times 3$ rotation matrix. In our degenerate configuration, the pose of the target related to the newly added image can be expressed as:
\begin{equation}
	\mathbf{T}_{pc}^{\prime} =\left[ {\begin{array}{*{20}{c}}
			{\mathbf{R}_{pc}\mathbf{R}_z}&- \mathbf{R}_{pc} \mathbf{R}_z\mathbf{t}_{cp}\\
			{\mathbf{0}_{1 \times 3}}&1
	\end{array}} \right],
	\label{eq:RbzRa}
\end{equation}
with 
\begin{equation}
	\mathbf{R}_z = \left[\begin{array}{ccc}
		\cos \theta & -\sin \theta & 0 \\
		\sin \theta & \cos \theta & 0 \\
		0 & 0 & 1
	\end{array}\right]
	\label{eq:Rz}
\end{equation}
where $\theta$ is the angle of rotation. Substituting \cref{eq:RbzRa,eq:Rz} into Eq. (6) in the paper, we get:
\begin{equation}
	\small
	\begin{aligned}
		\mathbf{h}_{1}^{\prime}&=\frac{\lambda^{\prime}}{\lambda}\left(\mathbf{h}_{1} \cos \theta+\mathbf{h}_{2} \sin \theta\right) \\
		\mathbf{h}_{2}^{\prime}&=\frac{\lambda^{\prime}}{\lambda}\left(-\mathbf{h}_{1} \sin \theta+\mathbf{h}_{2} \cos \theta\right)\\
		\mathbf{h}_{3}^{\prime}& = \frac{\lambda^{\prime}}{\lambda}\left[\mathbf{h}_{3}  +  (x  +  y \sin \theta  -  x \cos \theta) \mathbf{h}_{1} \right.\\
		&\left. \qquad \quad \quad  -  (y  -  x \sin \theta  -  y \cos \theta) \mathbf{h}_{2}
		\right]
	\end{aligned}
	\label{eq:ha123}
\end{equation}
Substituting \cref{eq:ha123} into constraint (8) in the paper, one of the constraints from $\mathbf{H}^{\prime} = [\mathbf{h}_{1}^{\prime} \quad \mathbf{h}_{2}^{\prime} \quad \mathbf{h}_{3}^{\prime}]$ can be expressed as:
\begin{equation}
	\begin{array}{r}
		\mathbf{h}_{1}^{\prime^{T}} \mathbf{K}^{-T} \mathbf{K}^{-1} \mathbf{h}_{2}^{\prime}=\frac{\lambda^{\prime}}{\lambda}\left[\cos 2\theta\left(\mathbf{h}_{1}^{\prime^{T}} \mathbf{K}^{-T} \mathbf{K}^{-1} \mathbf{h}_{2}\right)\right. \\
		\left.-\cos \theta \sin \theta\left(\mathbf{h}_{1}^{\prime^{T}} \mathbf{K}^{-T} \mathbf{K}^{-1} \mathbf{h}_{1}-\mathbf{h}_{2}^{\prime^{T}} \mathbf{K}^{-T} \mathbf{K}^{-1} \mathbf{h}_{2}\right)\right].
	\end{array}
	\label{eq:haKKha}
\end{equation}
Obviously, \cref{eq:haKKha} is a linear combination of the two constraints from $\mathbf{H} = [\mathbf{h}_{1} \quad \mathbf{h}_{2} \quad \mathbf{h}_{3}]$. Likewise,  we can deduce that the remaining constraints from $\mathbf{H}^\prime$ can be expressed as linear combinations of constraints provided by $\mathbf{H}$. Therefore, $\mathbf{H}^\prime$ can not provide additional constraints for solving the camera parameters.

\section{Performance of Initial Value Estimation}
\label{sec:PIVE}

In this experiment, we evaluate the performance of the algorithms in solving the initial value of camera linear parameters. The linear parameters of the camera include focal length and principal point. The generation of synthetic data is detailed in the paper. Note that the linear parameters do not include the distortion coefficients, so we set $d_1 = 0, d_2 = 0$.
\begin{figure}[htbp]
	\centering
	\includegraphics[width=0.3\linewidth]{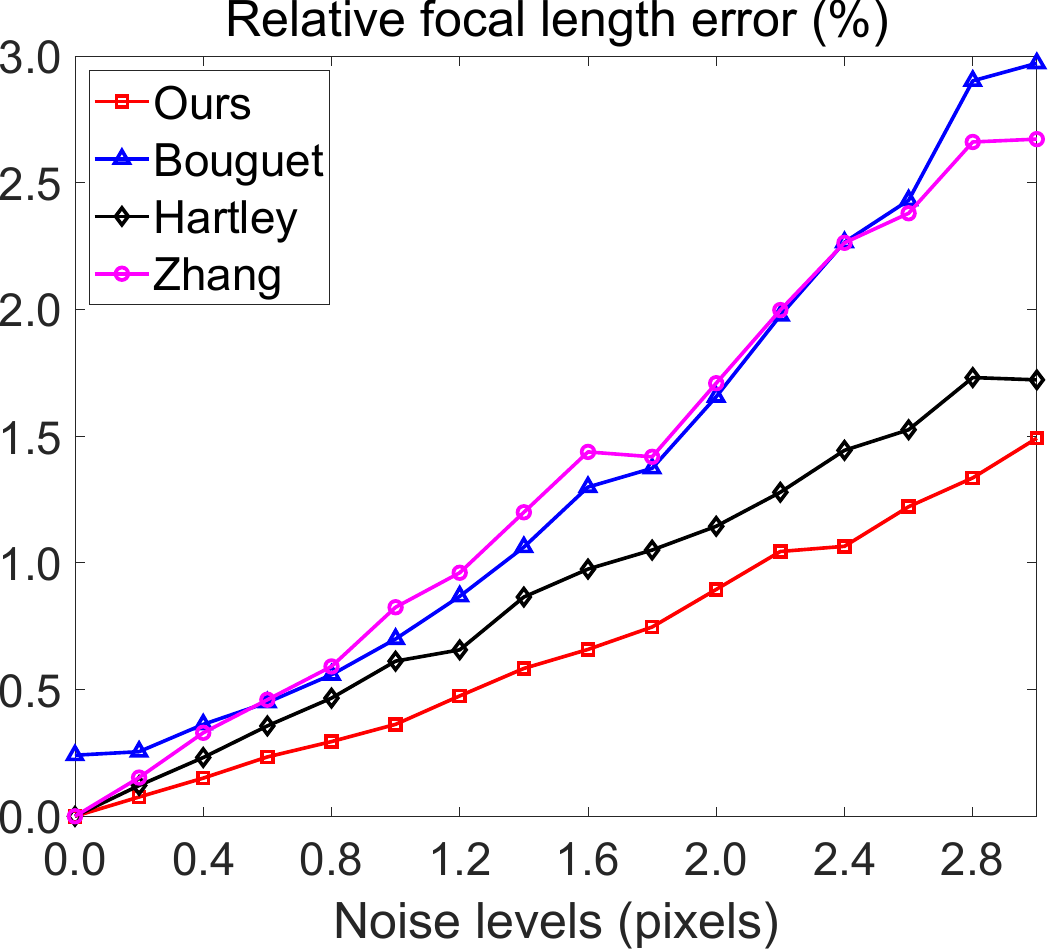}
	\quad
	\includegraphics[width=0.3\linewidth]{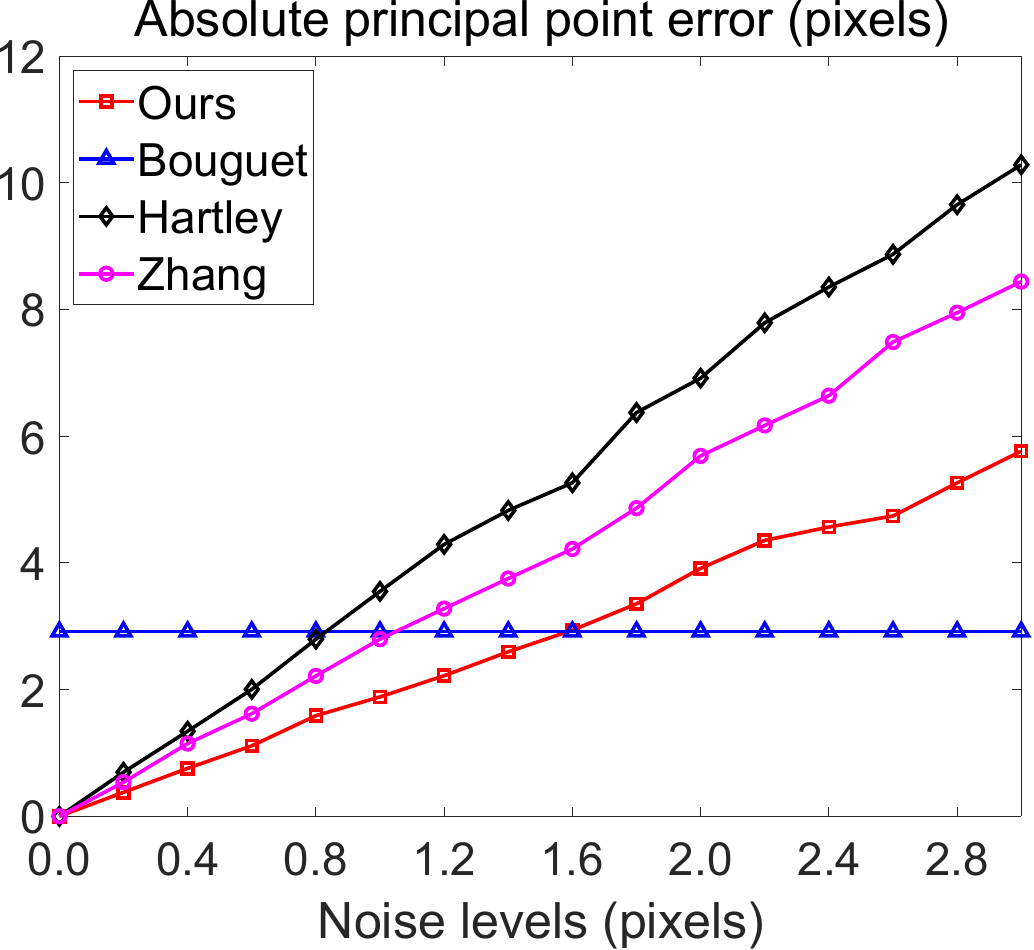}
	\caption{Robustness comparison of initial value estimation with increasing noise level. The relative focal length error (Left) and absolute principal point error (Right) increase linearly with the increase in noise. The proposed algorithm demonstrates greater robustness than other algorithms.}
	\label{fig:Init_Noise}
\end{figure}
\begin{figure}[htbp]
	\centering
	\includegraphics[width=0.31\linewidth]{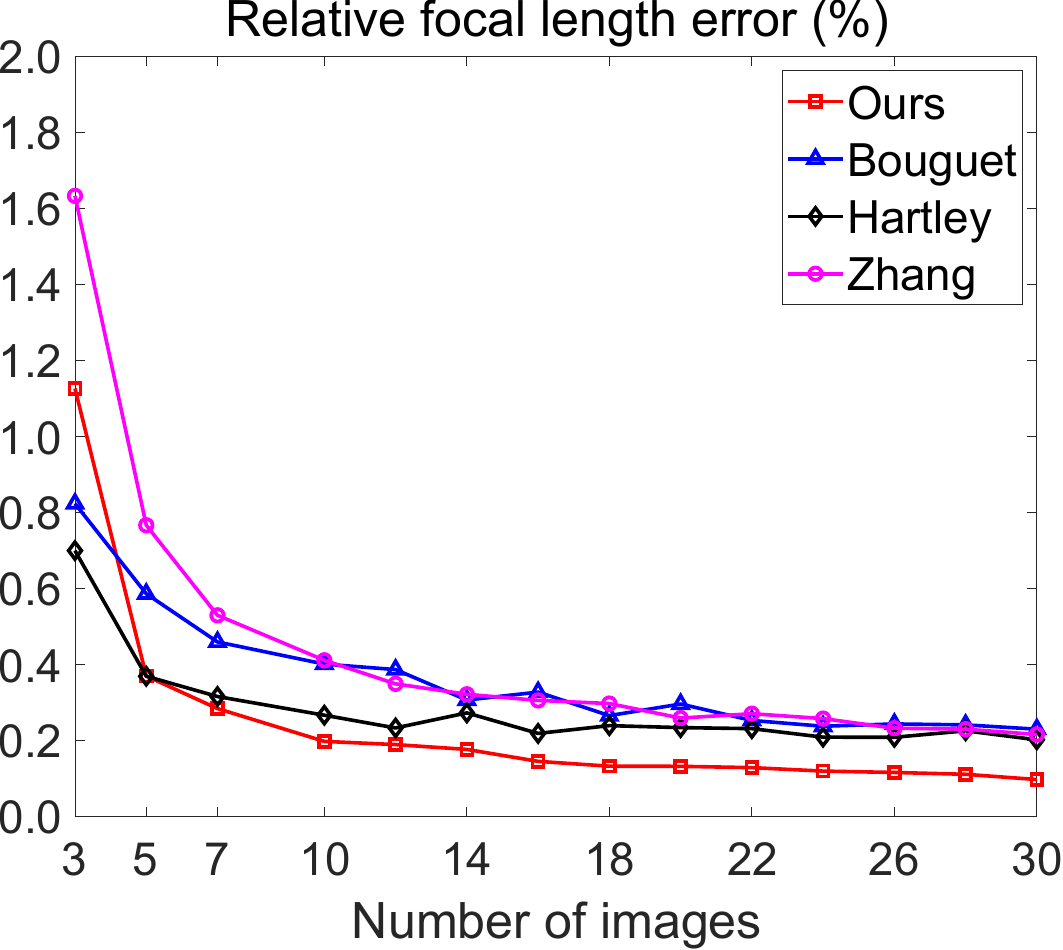}
	\quad
	\includegraphics[width=0.3\linewidth]{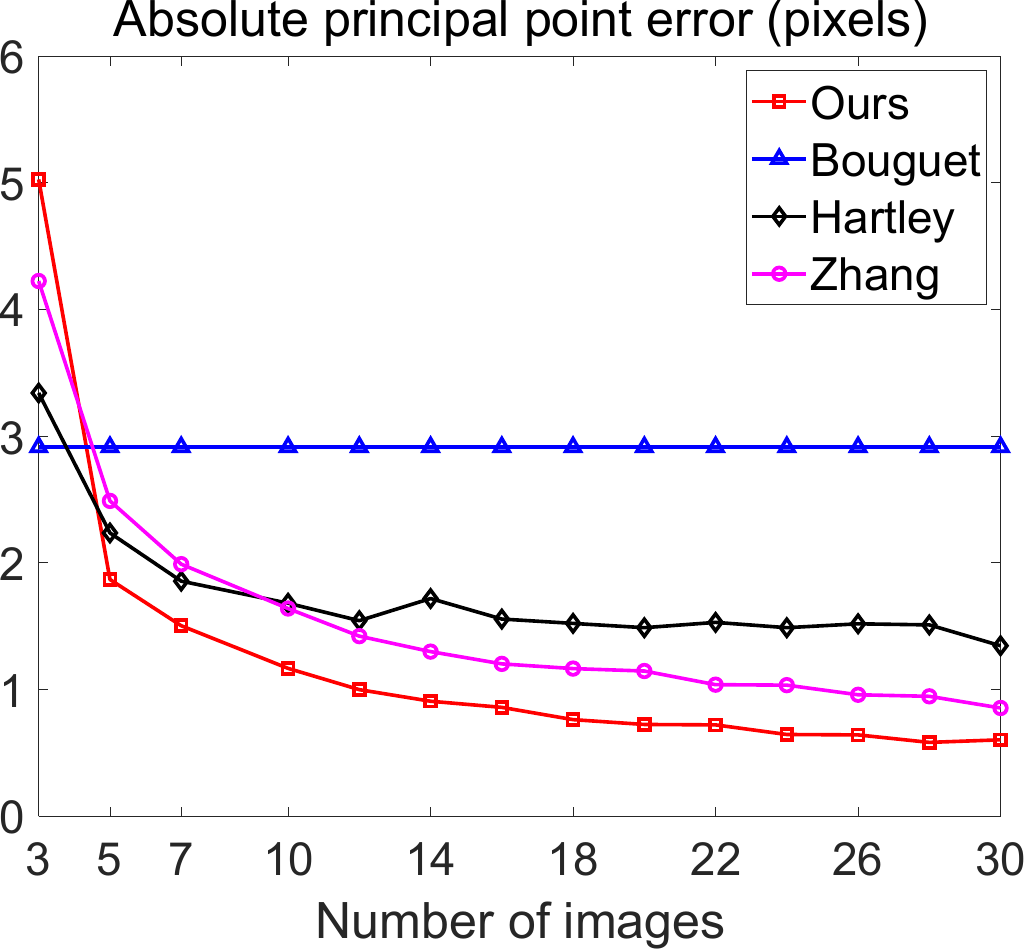}
	\caption{Accuracy comparisons of initial value estimation with the number of images increases. The statistical results of relative focal length error (Left) and absolute principal point error (Right) indicate that the calibration results become more stable with increased images. Our algorithm demonstrates superior accuracy compared to others.}
	\label{fig:Init_nImgs}	
\end{figure}

Firstly, we evaluate the performance of algorithms with respect to the noise levels. Gaussian noise with a mean of 0 and a standard deviation from 0 to 3.0 pixels is added to the image points. For each noise level, we used 15 images and performed 500 independent trials. As shown in \cref{fig:Init_Noise}, the calibration error exhibits a linear increase with the noise level. Our proposed algorithm demonstrates superior robustness against noise compared to the other three. At a noise level of 1 pixel, the initial focal length estimation error is less than $0.5\%$, and the initial principal point estimation error is less than 2.0 pixels. It should be noted that the calibration toolbox of \texttt{Bouguet} assumes the principal point to be fixed at the image center during the initial value estimation, which introduces a systematic bias. 

Secondly, we evaluate the performance of algorithms as the number of calibration images increases. The number of images used for calibration ranges from 3 to 30. 500 trials are conducted for each number, and the average results are shown. \cref{fig:Init_nImgs} demonstrates that the calibration error decreases as the number of images increases. When using 10 images for calibration, our focal length error is less than $0.2\%$, and the principal point error is around 1.0 pixels. When the number of images exceeds 5, our calibration algorithm consistently shows lower errors in estimating focal length and principal point.

In the following, we analyze the utilization of constraints in algorithms. \texttt{Zhang}' method effectively incorporates the homography constraint between the calibration targets and the images while disregarding the proposed spherical motion constraint. The calibration toolbox of \texttt{Bouguet} employs the same constraint as \texttt{Zhang}, but sets some default parameters that are reasonable priors in practice. The results depicted in \cref{fig:Init_Noise} and \cref{fig:Init_nImgs} demonstrate that \texttt{Bouguet} method shows a systematic bias in solving the initial value. \texttt{Hartley} fully utilizes the spherical motion constraint but overlooks the homography constraints between planes. However, our algorithm considers both spherical motion constraints and homography constraints, which makes ours perform better.

\section{Verification of Spherical Motion}
\label{sec:VSM}
\begin{figure}[htbp]
	\centering
	\includegraphics[width=0.3\linewidth]{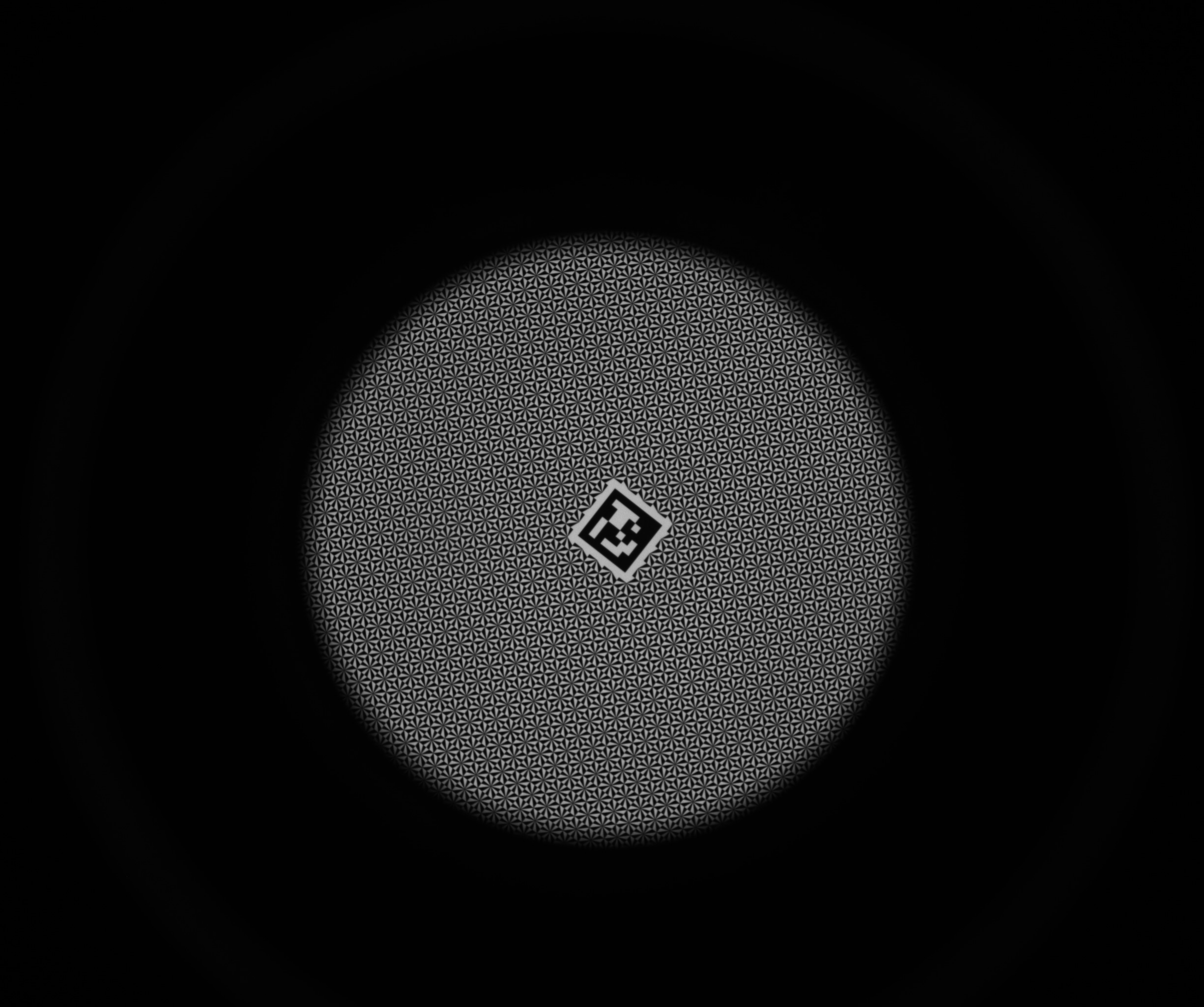}
	\quad
	\includegraphics[width=0.3\linewidth]{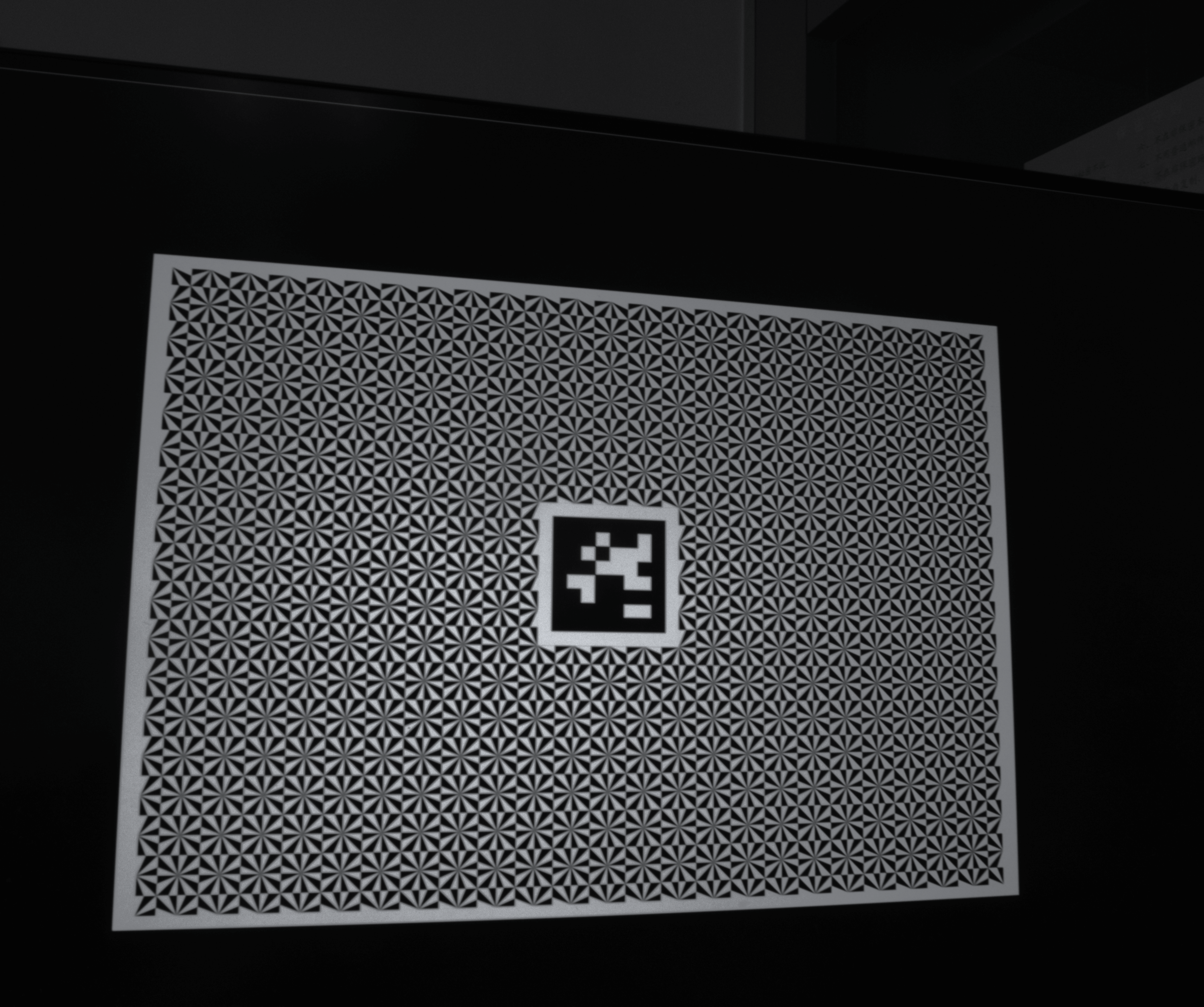}
	\caption{\textbf{Left:} Example of collimator image for pose estimation. \textbf{Right:} Example of monitor image for camera calibration.}
	\label{fig:Coll_moni}
\end{figure}

In Section 3.2 of the paper, we propose that the motion of the calibration target can be modeled as a spherical motion with respect to the fixed camera optical center through our collimator system. In this experiment, we verify the spherical motion model using real collimator images. A gray-scale camera with a resolution of $2448\times 2048$ and a lens with a focal length of 8$mm$ are employed to capture images from our collimator system. We collect a sequence of monitor images and estimate the camera parameters using calibration toolbox \cite{Bouguet2004}. Then, A total of 700 collimator images are used for verification. Given the calibrated camera parameters, the pose of all collimator images can be calculated using the perspective-n-points (P$n$P) \cite{OpenGV}. Examples of collimator and monitor images are shown in \cref{fig:Coll_moni}.  

The statistics on the position of 700 collimator images are presented in \cref{tab:CamPos}. The centroid of all positions is located at $t_{cp} = (94.5415, 137.3215, -252.3747)mm$. In a perfect spherical motion model, the positions of images in the calibration target coordinate system demonstrate consistency. The position of images may be slightly offset due to actual noise. The average radius of the spherical motion is $252.3747mm$. Notably, the standard deviation of the image positions is only ($0.4563\%, 0.4904\%, 0.1546\%$) relative to the spherical radius ($r = 252.3747mm$), which indicates a tightly concentrated distribution. In the synthetic data experiments of the paper, we show that our algorithm is superior to other algorithms when the imperfect spherical noise level is less than $2.143\%$ of the spherical radius. The parallelism of our collimator is obviously better than this noise level. The range represents the side length of the minimum enclosing cuboid that encompasses all the points. The variation range of the image positions relative to the spherical radius is ($2.0244\%, 2.0979\%, 0.6851\%$). In summary, it can be concluded that the image positions are nearly invariant. 
\begin{figure}[htbp]
	\centering
	\includegraphics[width=0.7\linewidth]{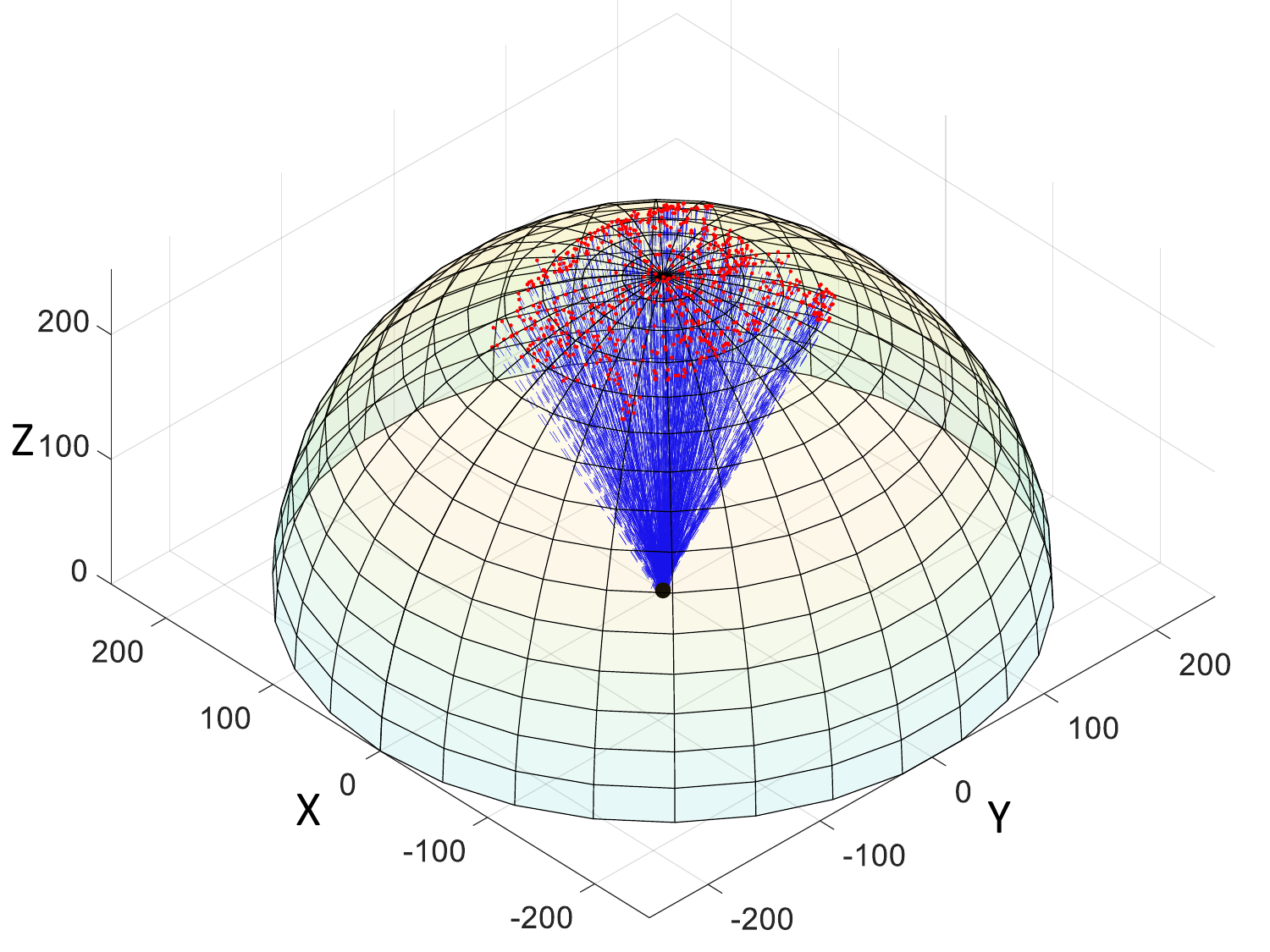}
	\caption{ Visualization of the pose of the calibration patterns. 
		The \textcolor{black}{black dot} represents the camera optical center located at the origin of the coordinate system. The \textcolor{red}{red dots} represent the positions of calibration targets, which fit nicely on the transparent sphere. The \textcolor{blue}{blue lines} depict the $z$-axis of the calibration patterns, converging towards the camera optical center, indicating that the calibration target's motion conforms to the spherical motion model.}
	\label{fig:Sphere}
\end{figure}

Moreover, we set the camera as the fixed reference coordinate system and determine the pose of all calibration targets. To visualize this, we generate a sphere with the camera's optical center as its center and the average distance as its radius ($r = 252.3747mm$). Subsequently, we depict the pose of the calibration targets on this sphere. The local coordinate system of the calibration targets is translated from the upper left corner to the $(94.5415, 137.3215, 0)mm$, but the directions remain unchanged. As depicted in \cref{fig:Sphere}, the position of the calibration targets (red dots) closely fits the spherical surface, with an average distance error of $0.3mm$. Additionally, the $z$-axis of calibration targets (blue lines) consistently aligns with the camera optical center (Black dot), which is consistent with Eq. (2) in the paper. The blue lines intersect with the $XY$-plane. The centroid of intersections is located at $(0.2850, 0.3754, 0)mm$, with a standard deviation of $(1.8734, 1.8077,0)mm$ . In summary, we can determine that the calibration target performs spherical motion with respect to the camera.
\begin{table}[tbp]
	\centering
	\caption{Statistical results of image position distribution. }
	\begin{tabular}{cccc}
		\toprule
		{} & x & y &z \\
		\midrule
		{\cellcolor{mypink}Mean$(mm)$ }& \cellcolor{mypink}94.5415 & \cellcolor{mypink}137.3215 & \cellcolor{mypink}-252.3747 \\
		
		\begin{tabular}{c}
			Standard \\Deviation$(mm)$
		\end{tabular}
		& 1.1516 & 1.2376 & 0.3902 \\
		\cellcolor{mypink}Range$(mm)$ & \cellcolor{mypink}5.1091 & \cellcolor{mypink}5.2945 & \cellcolor{mypink}1.7290\\
		\bottomrule
	\end{tabular}
	\label{tab:CamPos}
\end{table}

\section{Structure from Motion Results.}
\label{sec:SfM}
\begin{figure}[tbp]
	\centering
	\subfloat[Sand table]{
		\includegraphics[width=0.23\linewidth]{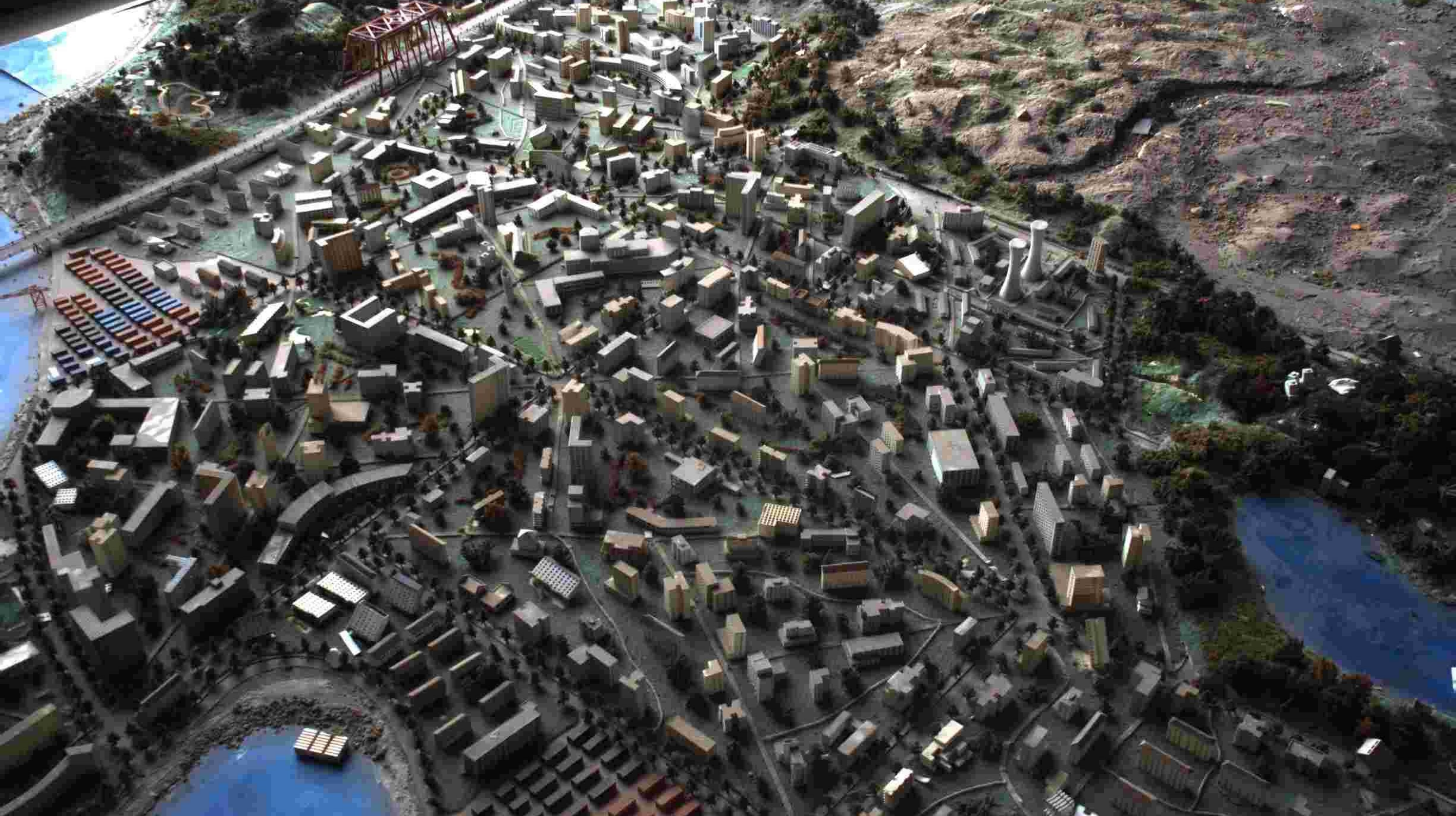} 
		\label{fig:sandTable}
	}
	\subfloat[Bouguet+Printed]{
		\includegraphics[width=0.25\linewidth]{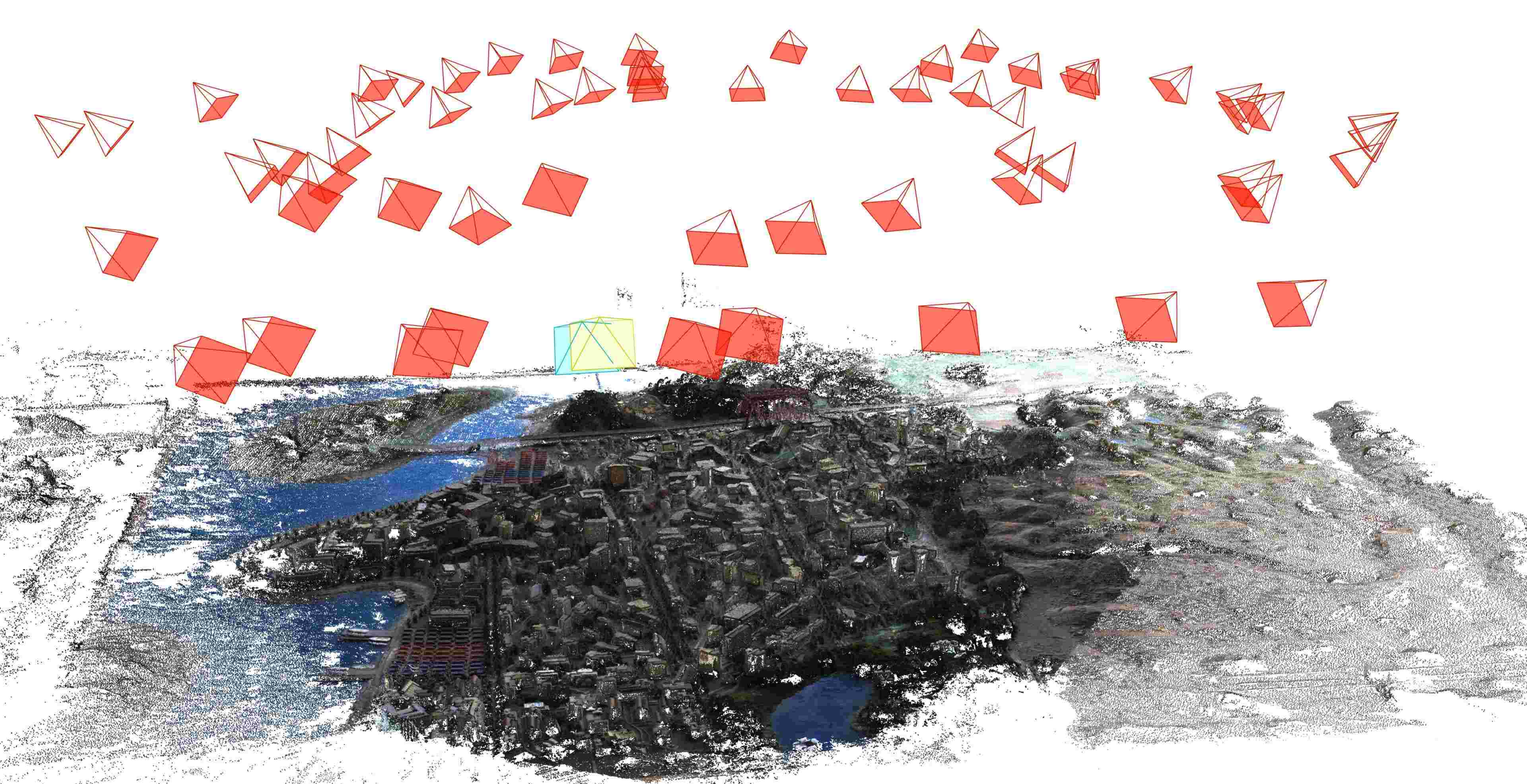}
		\label{fig:chessboard}
	}
	\subfloat[Zhang+Collimator]{
		\includegraphics[width=0.25\linewidth]{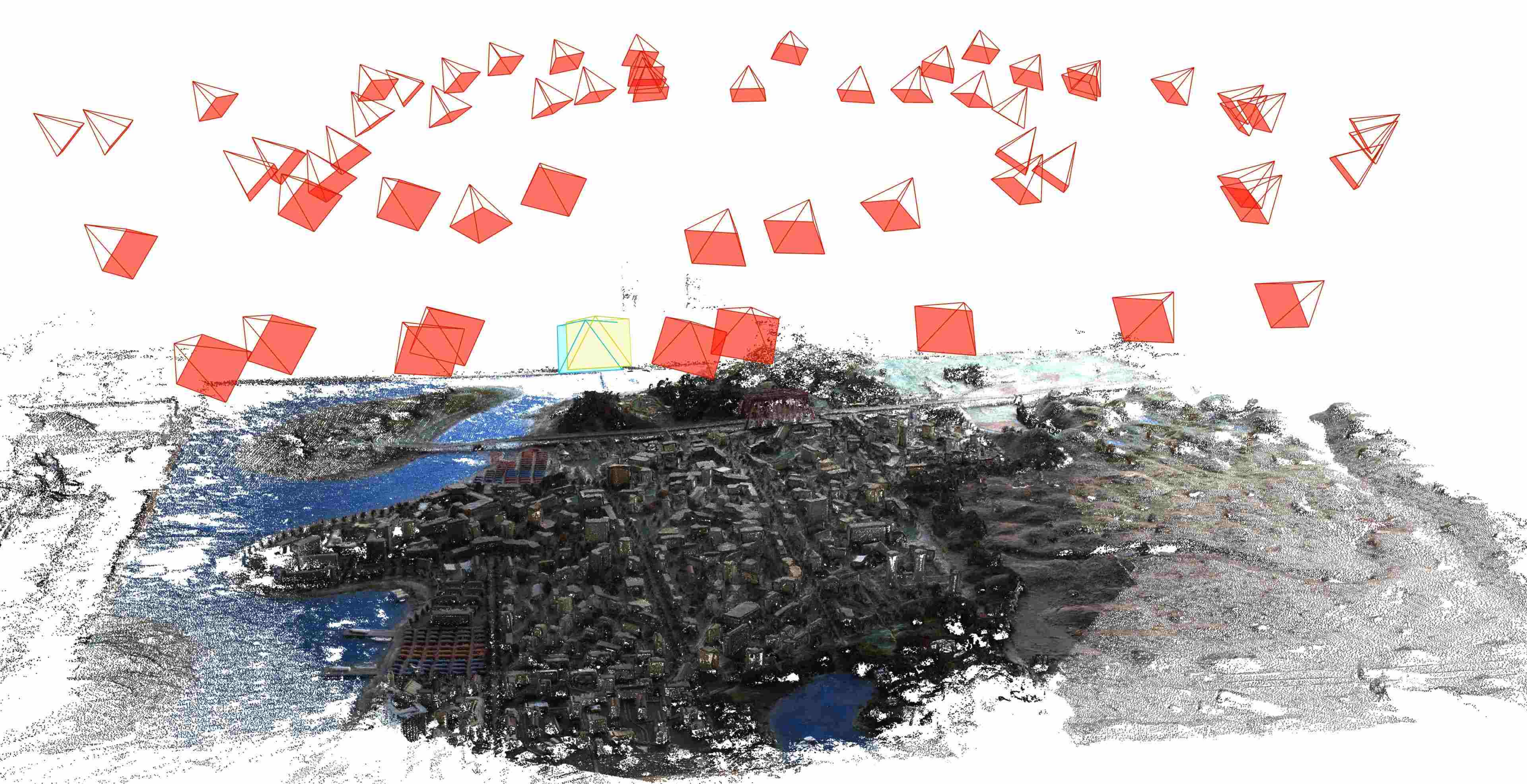}
		\label{fig:zhang_collimator}
	}	\\
	\subfloat[Bouguet+Collimator]{
		\includegraphics[width=0.25\linewidth]{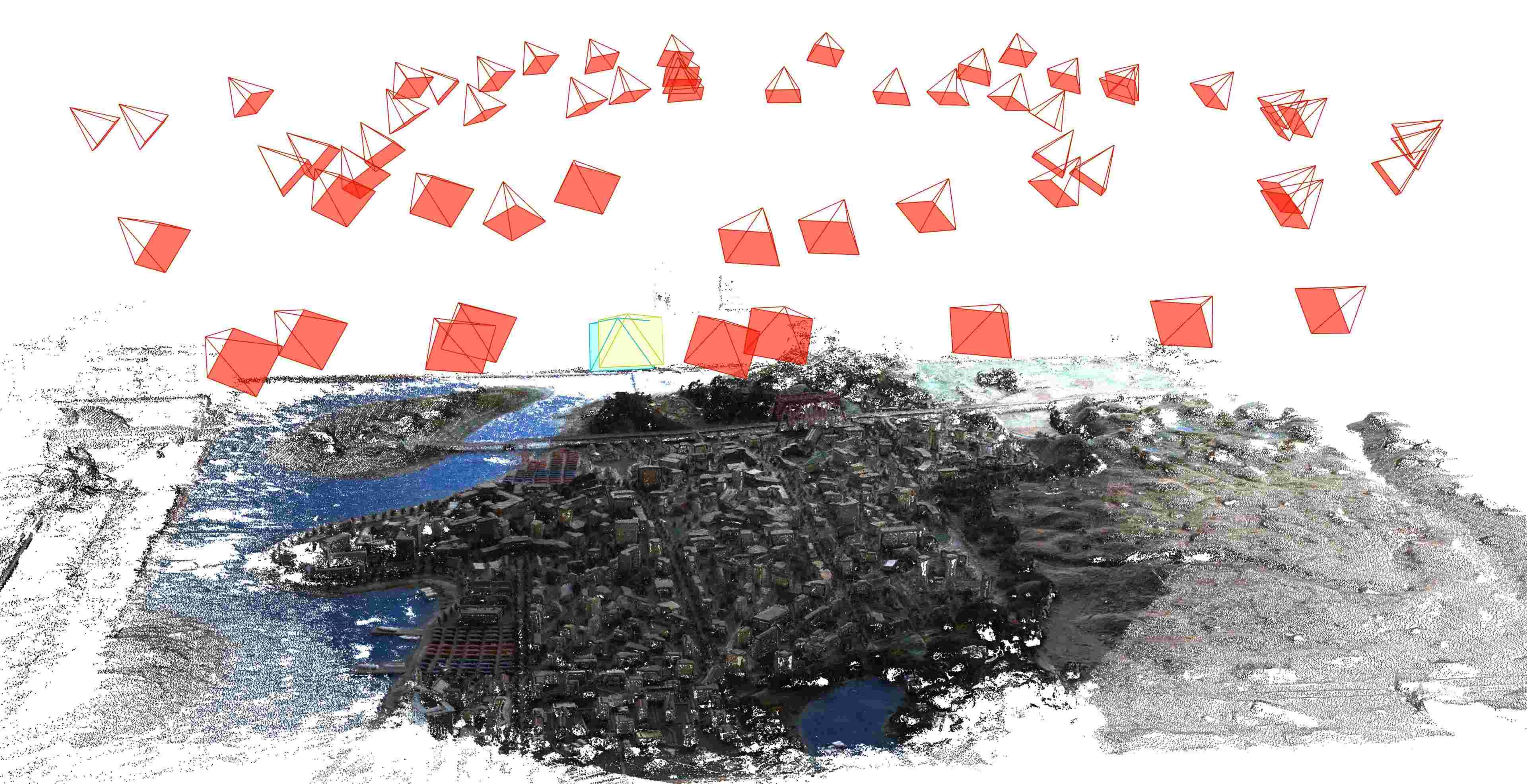}
		\label{fig:bouguet_collimator}
	}
	\subfloat[Hartley+Collimator]{
		\includegraphics[width=0.25\linewidth]{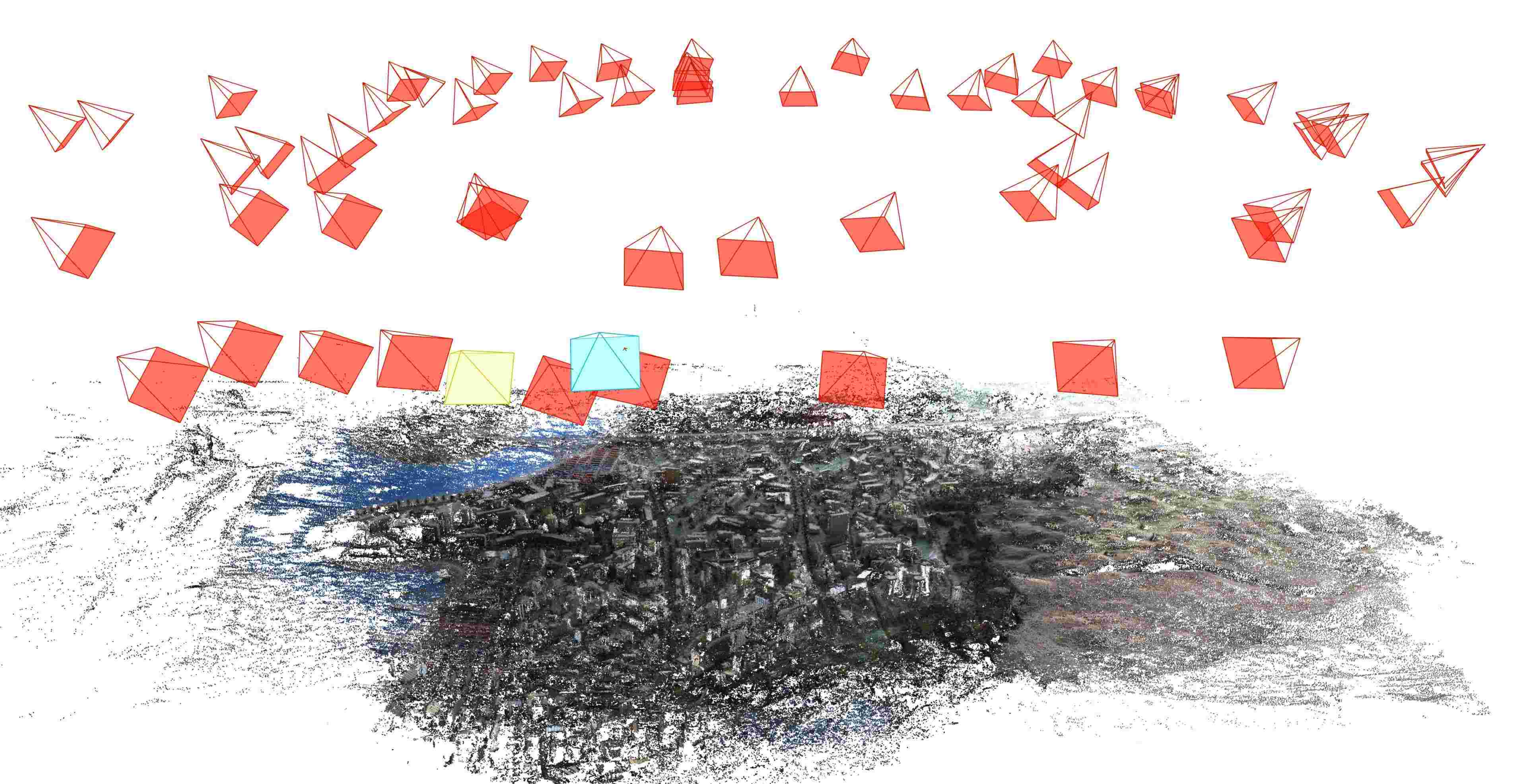}
		\label{fig:rotation_collimator}
	} 
	\subfloat[Ours+Collimator]{
		\includegraphics[width=0.25\linewidth]{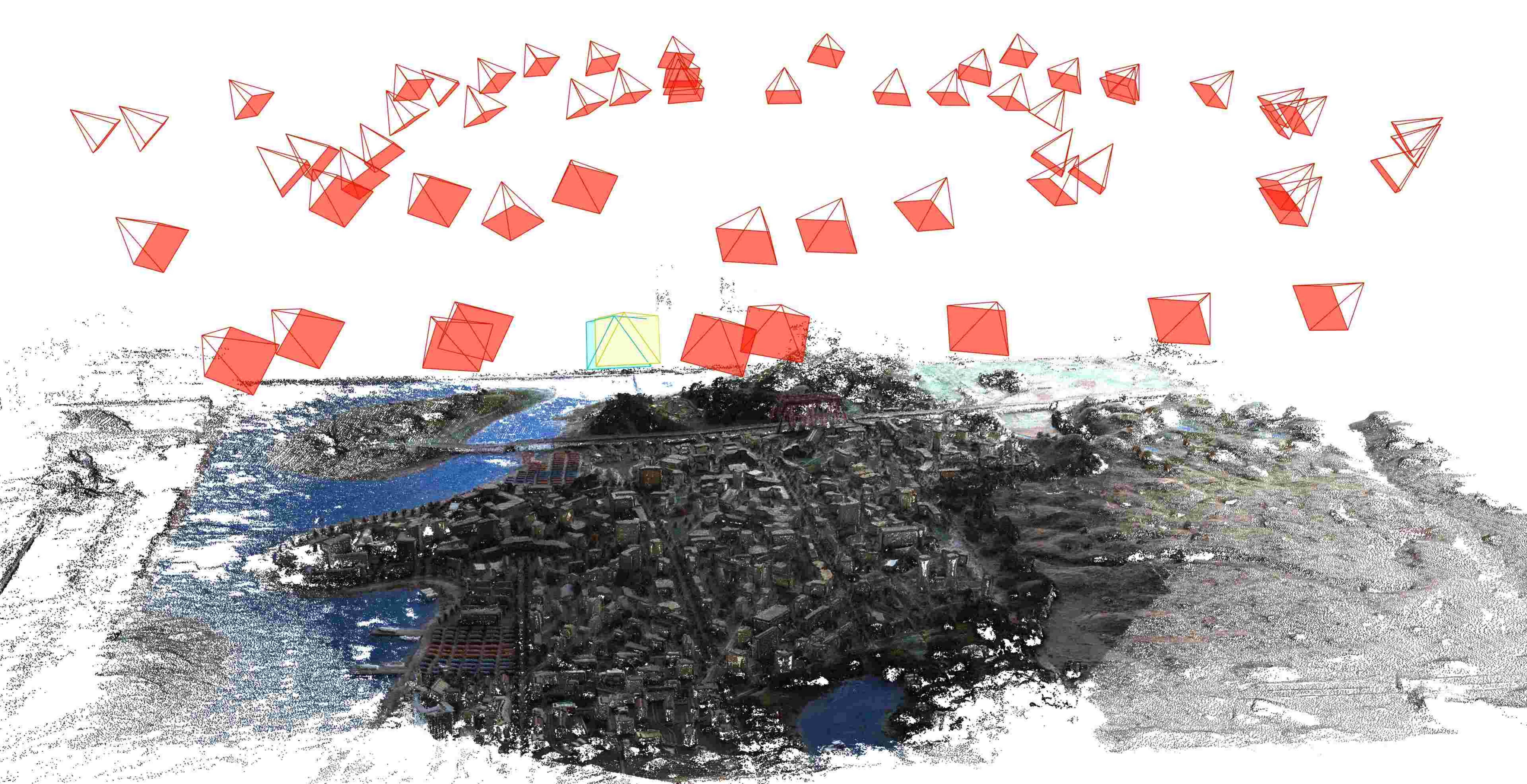}
		\label{fig:Ours_collimator}
	}
	\caption{Using COLMAP\cite{Colmap2016} to reconstruct an urban sand table.}
	\label{fig:SfM3D}
\end{figure}

In this experiment, we conduct a structure from motion (SfM) experiment to verify the accuracy and relevance of the proposed methods. The incremental SfM method COLMAP\cite{Colmap2016} is used to reconstruct an urban sand table, as shown in \cref{fig:sandTable}. Our experimental setup is similar to\cite{Peng2019}. We capture images around the sand table and duplicate the first image to the end of sequence. The camera parameters are obtained from 50 printed checkerboard and collimator images and served as input to the COLMAP\cite{Colmap2016}. The printed checkerboard images have multiple tilt angles to ensure accurate calibration. We set sequential matching between two adjacent images, reconstruction in image order, no loop detection, and no optimization of camera parameters. 
\begin{table}[tbp]
	\centering
	\caption{Comparison of SfM results}
	\resizebox{0.7\linewidth}{!}{
		\begin{tabular}{c|c|cccc}
			\hline
			\multirow{2}{*}{}&{Printed}&\multicolumn{3}{c}{Collimator}&{} \\
			\hline
			{}&{Bouguet\cite{Bouguet2004}}&{Zhang\cite{Zhang2000}}&{Bouguet\cite{Bouguet2004}}&{Hartley\cite{Hartley1997}}&{Ours}\\
			\hline
			\cellcolor{mypink}{$\varepsilon_\mathbf{R}$}&\cellcolor{mypink}{0.9821}&\cellcolor{mypink}{0.5012}&\cellcolor{mypink}{0.5049}&\cellcolor{mypink}{8.2561}&\cellcolor{mypink}{\textbf{0.3971}}\\
			{$\varepsilon_\mathbf{t}$}&{0.0057}&{0.0032}&{0.0033}&{0.0317}&{\textbf{0.0029}}\\
			\cellcolor{mypink}{$\varepsilon_{proj}$}&\cellcolor{mypink}{18.6869}&\cellcolor{mypink}{11.2434}&\cellcolor{mypink}{10.7638}&\cellcolor{mypink}{150.5852}&\cellcolor{mypink}{\textbf{10.1284}}\\
			{Point\#}&{{34412}}&{34395}&{34109}&{34340}&\textbf{34858}\\
			\cellcolor{mypink}{$\varepsilon_{SfM}$}&\cellcolor{mypink}{0.6121}&\cellcolor{mypink}{0.5989}&\cellcolor{mypink}{0.6007}&\cellcolor{mypink}{0.8536}&\cellcolor{mypink}{\textbf{0.5776}}\\
			\hline
		\end{tabular}
	}
	\label{tab:SfM}
\end{table}

For a more accurate calibration, the pose deviation between the first and last image is smaller. The pose deviation is measured by rotation error $\varepsilon_{\mathbf{R}}$, translation error $\varepsilon_{\mathbf{t}}$, and re-projection error $\varepsilon_{proj}$ as referenced in\cite{Peng2019}. \cref{fig:SfM3D} shows dense reconstruction results and the pose deviations of the first and last images (marked as cyan and yellow, respectively). Intuitively, the result from our method shows a higher level of alignment between the two images. Quantitative results are shown in \cref{tab:SfM}. Our method exhibits minimum pose deviation and maintains a minimum average re-projection error $\varepsilon_{SfM}$ in the SfM process. The SfM outcomes indicate improved accuracy in the collimator image and highlight the beneficial impact of the spherical constraint on accuracy. 

\end{document}